\documentclass[letterpaper]{article} %
\usepackage{aaai25}  %
\usepackage{times}  %
\usepackage{helvet}  %
\usepackage{courier}  %
\usepackage[hyphens]{url}  %
\usepackage{graphicx} %
\usepackage{natbib}  %
\usepackage{caption} %
\usepackage{algorithm}
\usepackage{algorithmic}
\usepackage{dsfont}

\usepackage[T1]{fontenc}
\usepackage{xcolor}
\usepackage{booktabs}
\usepackage{multirow, multicol}
\usepackage{subcaption}
\usepackage{url}
\usepackage{mathtools}
\usepackage{yhmath}
\usepackage{amsthm}
\usepackage{pgfplots}

\pgfplotsset{compat=1.18}
\usepackage{amsfonts}
\usetikzlibrary{patterns}
\renewcommand{\hat}{\widehat}

\newcommand{\Itrain}{\mathcal I_{\textrm{train}}}
\newcommand{\Ical}{\mathcal I_{\textrm{cal}}}
\newcommand{\Itest}{\mathcal I_{\textrm{test}}}
\newcommand{\Scal}{S^{\textrm{cal}}}
\newcommand{\Stest}{S^{\textrm{test}}}
\newcommand{\rcal}{r^{\textrm{cal}}}
\newcommand{\rtest}{r^{\textrm{test}}}
\newcommand{\scal}{s^{\textrm{cal}}}
\newcommand{\stest}{s^{\textrm{test}}}

\newtheorem{prop}{Proposition}

\definecolor{matblue}{HTML}{1F77B4}
\newtheorem{theorem}{Theorem}
\newtheorem*{remark}{Remark}

\newcounter{protocol}%
\newcounter{algorithm saved}%


\usepackage{adjustbox}
\usepackage{enumitem}

\usepackage{newfloat}
\usepackage{listings}
\DeclareCaptionStyle{ruled}{labelfont=normalfont,labelsep=colon,strut=off} %
\floatstyle{ruled}
\newfloat{listing}{tb}{lst}{}
\floatname{listing}{Listing}
\title{Conformalized Interval Arithmetic with Symmetric Calibration}
\author{
    Rui Luo\textsuperscript{\rm 1}\thanks{Corresponding author}  and
    Zhixin Zhou\textsuperscript{\rm 2}
}
\affiliations{
    \textsuperscript{\rm 1}Department of Systems Engineering, City University of Hong Kong, Hong Kong SAR, China\\
    \textsuperscript{\rm 2}Alpha Benito Research, Los Angeles, USA\\
    \texttt{ruiluo@cityu.edu.hk, zzhou@alphabenito.com}
}

\usepackage{bibentry}

\begin{document}

\maketitle

\begin{abstract}
Uncertainty quantification is essential in decision-making, especially when joint distributions of random variables are involved. While conformal prediction provides distribution-free prediction sets with valid coverage guarantees, it traditionally focuses on single predictions. This paper introduces novel conformal prediction methods for estimating the sum or average of unknown labels over specific index sets. We develop conformal prediction intervals for single target to the prediction interval for sum of multiple targets.  Under permutation invariant assumptions, we prove the validity of our proposed method. We also apply our algorithms on class average estimation and path cost prediction tasks, and we show that our method outperforms existing conformalized approaches as well as non-conformal approaches. 
\end{abstract}

\begin{links}
    \link{Code}{https://github.com/luo-lorry/CIA}
    \link{Extended version}{https://arxiv.org/abs/2408.10939}
\end{links}

\section{Introduction}

Conformal prediction~\cite{vovk2005algorithmic} has gained popularity as a method for uncertainty quantification due to its validity guarantee. It plays a crucial role in modern decision-making processes, particularly in domains where understanding the joint distribution of random variables is essential. Under the exchangeability assumptions on the calibration and test samples, conformal prediction can provide prediction sets with valid coverage guarantees. Existing works have studied prediction intervals for single targets~\cite{lei2018distribution, romano2019conformalized, sesia2021conformal}. Given a black box algorithm to estimate the target label, current conformalized algorithms can provide prediction intervals for the target with reliable coverage probability. However, although numerous studies have developed the methodology of conformal prediction, they are limited to predicting a single label, leaving a gap in finding the prediction set for the sum of labels.

This gap is particularly relevant in applications such as transductive conformal prediction on traffic networks. For example, existing Graph Neural Network (GNN) methods can predict the label of each road, where the label can be considered as the cost of traversing that road. This problem has been studied in~\cite{huang2024uncertainty, zargarbashi2023conformal, zhao2024conformalized}. Conformalized GNN can output a prediction set of each edge's label, but the cost of a route, which is the sum of the labels of the edges on the route, cannot be directly obtained by applying conformal prediction to individual edges. This challenge arises from two main aspects. First, the sum or average of random variables involves the convolution of the density function, and simple interval arithmetic, such as adding up the lower and upper bounds, cannot provide a confidence interval with the desired coverage. Second, the coverage of each confidence interval is in a marginal sense, so the coverage of two labels from two confidence intervals are dependent events. Consequently, it is difficult to use the conformal prediction set for a single label to devise a prediction set for multiple labels. To address this critical gap in the current literature on conformal prediction and expand its applicability to a broader range of uncertainty quantification problems, we introduce the method \emph{Conformal Interval Arithmetic (CIA)}, specifically designed to estimate the average or other symmetric functions of unknown labels over a certain index set, demonstrating the usefulness of our problem setting in many applications where people are interested in obtaining estimates about multiple labels.

Our main proposed method can be described as follows. The core idea is to establish a confidence interval using the exchangeability of the groups of indices. This method involves finding the absolute value of the difference between the sum of labels in a group and its prediction, or absolute residual, and using this as the score function for split conformal prediction. Assuming the groups of indices are exchangeable, we can provide prediction sets with valid marginal coverage. However, this approach cannot handle the case when the calibration samples and the test samples belong to different groups. To address this issue, we introduce a new split conformal prediction method called \emph{symmetric calibration}, which creates a scenario in which the calibration set and the test set play symmetric roles. At the group level, each index within the same group has equal chance of being assigned to either a calibration sample or a test sample. This ensures that the residual from the sums of calibration samples and the sums of test samples have identical distributions. By leveraging the exchangeability of the groups of samples in the calibration set, we can devise a conformal prediction set for the sum of residuals of calibration samples. Since both the calibration samples and the test samples share the same distribution, this conformal prediction set is also valid for the sum of the test samples.

In addition to the methodology of symmetric calibration and the validity guarantee discussed above, we will introduce some important extensions of our method to make it more widely applicable and efficient. First, symmetric calibration is easy to understand when the groups of indices are disjoint. We relax this condition to allow for small amounts of overlap between groups. Second, given prior knowledge of the number of unknown labels being summed, we can stratify the prediction into different levels. Finally, we can incorporate other conformal prediction methods, such as conformalized quantile regression~\cite{romano2019conformalized}, to improve prediction efficiency.

To summarize the contribution of this paper:
\begin{enumerate}
    \item We introduce the idea of conformalized interval arithmetic (CIA) to provide confidence interval for sum of multiple unknown labels.
    \item We develop the technique of symmetric calibration to address the issue that unknown label can appear in different groups. The proposed method has valid coverage guarantee theoretically and empirically. 
    \item We apply our methods on datasets in~\cite{sesia2020comparison}, and show the reliability of our method. The code of our method will be published as open source. 
\end{enumerate}

\section{Preliminary and Problem Setup}

We will first provide an overview of conformal prediction in the literature. Then, we will study conformalized interval arithmetic (CIA) in a simple setting, where the calibration set and test set are exchangeable at the group level. In this case, we can use the residuals of the sums to perform split conformal prediction on the test group.

\subsection{Split Conformal Prediction}\label{sec:conformal:prediction}

Conformal prediction has been applied to the following types of problems. Suppose a dataset with an index set $\mathcal{I}$ is partitioned into training, calibration, and test sets, with index sets $\Itrain$, $\Ical$, and $\Itest$, respectively. Firstly, we apply a black-box model to the training data $\{(x_i, y_i)\}_{i \in \Itrain}$ and obtain a function $\hat{f}$ such that $\hat{y}_i = \hat{f}(x_i)$. We define a conformity score function $s(x, y)$, which depends on $\hat{f}$. A larger value of the score function indicates that $y$ is less likely to be a true label. A typical choice of the score function is
\begin{align*}
    s_i=s(x_i,y_i)= |y_i-\hat y_i|.
\end{align*}
We will first focus on this choice and then extend it to other choices in later sections. For data in $\{(x_i, y_i)\}_{i \in \Ical}$, we evaluate $s_i=|y_i-\hat y_i|$ and let $Q_{1-\alpha}$ be the $\lceil(1 + |\Ical|)(1 - \alpha)\rceil$-th smallest value of $|y_i-\hat y_i|$ for $i\in\Ical$\footnote{If $\lceil(1 + |\Ical|)(1 - \alpha)\rceil > |\Ical|$, then $Q_{1-\alpha} = \infty$. Here, $Q_{1-\alpha}$ represents the quantile of the scores. The meaning of $Q_{1-\alpha}$ may vary depending on how the scores are defined in different sections.}. Then, for $i \in \Itest$, the prediction set of $y_i$ is defined as
\begin{align*}
\mathcal{C}(i) = \{y:|y-\hat y_i| \le Q_{1-\alpha}\} 
= [\hat y_i-Q_{1-\alpha}, \hat y_i+Q_{1-\alpha}].
\end{align*}
Under the assumption of exchangeability of the samples in the calibration set and the test set, it can be shown that this conformal prediction set has coverage of at least $1-\alpha$. In prediction tasks, the resulting prediction set $\mathcal C(i)$ is usually an interval. Our present work will focus on the generalization of these prediction intervals.

\subsection{Problem Setup and a Naive Approach}\label{sec:naive}

In many applications, we are not just satisfied with the prediction set for a single label $y_i$. Unlike existing works, this paper aims to provide the prediction set of the sum of variables, i.e., a prediction set $\mathcal{C}({S})$ for $\sum_{i \in {S}} y_i$. We will apply conformal prediction techniques so that $1-\alpha$ coverage is guaranteed under certain exchangeability assumption.

The choice of $S\subseteq \Itest$ can depend on fixed or random procedures. Some concrete examples will be considered in Section~\ref{sec:exp}. We can start with a simple example. Let $S=\{i,j\}\subseteq \Itest$. Suppose a simple split conformal prediction provides prediction intervals $\mathbb P(Y_i\in[L_i,U_i])\ge 1-\alpha$ and $\mathbb P(Y_j\in[L_j,U_j])\ge 1-\alpha$, then the addition operation of interval arithmetic gives
\begin{align*}
    &\mathbb P(Y_i+Y_j\in [L_i+L_j, U_i+U_j])\\
    &\ge \mathbb P(Y_i\in[L_i,U_i] \text{ and } Y_j\in[L_j,U_j])\\
    &\ge 1-2\alpha. 
\end{align*}
Hence, the addition operation of interval arithmetic of $1-\alpha$ confidence intervals cannot guarantee $1-\alpha$ coverage. One possible correction is to first find $1-\alpha/2$ confidence intervals for $Y_i$ and $Y_j$, then construct the prediction interval by adding the intervals. This can be generalized using Bonferroni correction to predict the sum of any number of unknown labels, without requiring assumptions about the dependencies between variables. However, this method clearly lacks efficiency. Our method will be compared with Bonferroni correction in empirical experiments in Section~\ref{sec:exp}.

\section{Conformal Interval Arithmetic} \label{sec:cia}

We first study a simple case, in which the problem can be reduced to existing conformal prediction techniques. Suppose we have disjoint index sets $S_1, S_2, \dots, S_K, S_{K+1}$ and suppose we observe samples $(x_i, y_i)$ for $i\in S_1\cup S_2\cup\dots\cup S_K$, and $x_j$ for $j\in S_{K+1}$ as test samples. We can then calculate the residual as the score function on the set level:
\begin{align*}
    s_k := s(z_k) := \left| \sum_{i\in S_k} (y_i-\hat y_i) \right|, k\in [K]
\end{align*}
where $K=\{1,\dots,K\}$ and $z_k = (x_i,y_i)_{i\in S_k}$ denotes a group of pairs of $(x,y)$. The following procedure follows similarly from the conformal prediction from the previous section. Let $Q_{1-\alpha}$ be the $\lceil(1+K)(1-\alpha)\rceil$-th smallest value of $s_k, k\in[K]$. We define the prediction set $\mathcal C(S_{K+1})$ as
\begin{align}\label{eq:pred:simple}
    \left[ \sum_{i\in S_{K+1}} \hat y_i-Q_{1-\alpha}, \sum_{i\in S_{K+1}} \hat y_i+Q_{1-\alpha} \right].
\end{align}
This procedure provides conformal prediction for the sum of unknown labels and is called Conformal Interval Arithmetic (CIA) in this paper. Assuming group exchangeability, it can be guaranteed that the coverage probability is valid.

\begin{prop}\label{prop:simple}
    Suppose that $Z_k$ is group exchangeable in the sense of for any permutation $\pi$ on $[K+1]$: 
    \begin{align}\label{eq:group:exchange}
        \mathbb P(Z_1, \dots, Z_{K+1}) = \mathbb P(Z_{\pi(1)}, \dots, Z_{\pi(K+1)}), 
    \end{align}
    then the prediction set $\mathcal C(S_{K+1})$ in~\eqref{eq:pred:simple} satisfies
    \begin{align*}
        \mathbb P\left(\sum_{i\in S_{K+1}}y_i\in \mathcal C(S_{K+1})\right)\ge 1-\alpha. 
    \end{align*}
\end{prop}

The idea of conformal interval arithmetic can be applied to the simple case where the test sample with unknown labels belongs only to the same group, i.e., $S_{K+1}$, and $y_i$ for $i\in S_1, \dots, S_K$ must be fully observed. However, this assumption may not hold for many scenarios, such as cost estimation of edges in a route, as discussed in the introduction. Consider a traffic network where we randomly sample two nodes and find the shortest path between them. If a sufficient amount of edge weights are not observed, then only a few routes will contain all edges with observed weights. Let $S_1, \dots, S_K$ be the set of edges on a path, determined by the network structure and edge costs. For unobserved weights, they can be estimated by their predictions. In this method, every shortest path can contain edges with unobserved weights, and edges can belong to multiple paths. These issues make the simple case method inapplicable.

\section{Symmetric Calibration}

Let us formally state the problem. We aim to provide a prediction set for $\sum_{i\in S_k} y_i$. Since $y_i$ is not observed only for $i\in\mathcal I$, it is equivalent to finding the prediction set for $\sum_{i\in S_k\cap \Itest} y_i$. In the previous section, we considered a simple case where $S_{K+1}=\Itest$. Here, we will consider a general case where every $S_k$ can contain indices of unobserved labels. To tackle this problem, we introduce the method of \emph{symmetric calibration}.

Before training the model $\hat f$, we first hold out a calibration set that has the same size as the test set, i.e., $|\Ical|=|\Itest|$. Note that we assume the number of observed labels must be greater than the number of unobserved labels. Now, we focus on the calibration and test samples. Let us assume
\begin{align*}
    S_1\cup\dots\cup S_{K}\cup S_{K+1} = \Ical\cup\Itest,
\end{align*}
and let us define
\begin{align}\label{eq:score:cia}
    \Scal_k := S_k\cap\Ical
    \quad\text{ and }\quad
    \Stest_k = S_k\cap\Itest.
\end{align}
Our task becomes finding the prediction set for $\sum_{i\in\Stest_k} y_i$ for every fixed $k\in[K+1]$. To simplify the presentation, let us focus on the prediction set of $\sum_{i\in\Stest_{K+1}} y_i$. Other prediction sets and the coverage guarantee in the theorems can be obtained in the same manner. Using conformal interval arithmetic, we define the score function similar to~\eqref{eq:score:cia}:
\begin{align}\label{eq:split:score}
    \scal_k = \left|\sum_{i\in \Scal_k} (y_i-\hat y_i)\right|
\end{align}
Let $Q_{1-\alpha}$ be the $\lceil(1+K)(1-\alpha)\rceil$-th smallest value of $\scal_k, k=1,\dots, K$. Then we define the prediction set for $\sum_{i\in \Stest_{k}}y_i$, denoted by $\mathcal C(\Stest_{k})$:
\begin{align}\label{eq:pred:symmetric}
    \left[ \sum_{i\in \Stest_{k}} \hat y_i-Q_{1-\alpha}, \sum_{i\in \Stest_{k}} \hat y_i+Q_{1-\alpha} \right].
\end{align}
The above procedure is summarized in Algorithm~\ref{alg:cia}.
\begin{algorithm}
\caption{CIA with Symmetric Calibration}
\begin{algorithmic}[1]
\STATE \textbf{Input:} labeled samples $\{(x_i, y_i)\}_{i\in \mathcal I\backslash\Itest}$, unlabeled samples $\{x_i\}_{i\in \Itest}$, a black box algorithm $\mathcal B$, disjoint index sets $S_1, \dots, S_K\subseteq \mathcal I$.  
\STATE \textbf{Output:} Prediction sets for $\sum_{i\in \Stest_{K+1}} y_i$. 
\STATE Hold out a calibration set $\Ical$ from $\mathcal I\backslash\Itest$ such that $\Ical$ has the same size as $\Itest$. 
\STATE $\hat f\leftarrow \mathcal B(\{(x_i,y_i)\}_{i\in \mathcal I\backslash(\Ical\cup\Itest)}$.
\STATE $\hat y_i\leftarrow \hat f(x_i)$ for $i\in\Ical\cup\Itest$. 
\STATE $\scal_k = \left|\sum_{i\in \Scal_k} (y_i-\hat{f}(x_i))\right|$ for $k\in[K]$
\STATE $Q_{1-\alpha}\leftarrow\lceil(1+K)(1-\alpha)\rceil$-th smallest value of $\{\scal_k:k\in[K]\}\cup\{\infty\}$.
\STATE $\mathcal C(\Stest_{K+1})\leftarrow \left[ \sum_{i\in \Stest_{K+1}} \hat y_i-Q_{1-\alpha}, \sum_{i\in \Stest_{K+1}} \hat y_i+Q_{1-\alpha} \right].$
\STATE \textbf{Output:} $\mathcal C(\Stest_{K+1})$.
\end{algorithmic}\label{alg:cia}
\end{algorithm}

We will analyze the coverage probability of $\mathcal C(\Stest_{K+1})$. In this section, we assume that $S_1, \dots, S_K, S_{K+1}$ are disjoint. We will introduce the reasoning behind this in two steps. Firstly, under the assumption of group exchangeability, the tuple of score functions $(\scal_1, \dots, \scal_K, \scal_{K+1})$ is exchangeable. $\scal_{k}$ has a uniform rank over $[K+1]$, so $\scal_{K+1}\le Q_{1-\alpha}$ with probability at least $1-\alpha$. Equivalently, this suggests that the interval
\begin{align}\label{eq:pred:symmetric:test}
\left[ \sum_{i\in \Scal_{K+1}} \hat y_i-Q_{1-\alpha}, \sum_{i\in \Scal_{K+1}} \hat y_i+Q_{1-\alpha} \right]
\end{align}
covers $\sum_{i\in \Scal_{K+1}} y_i$ with probability at least $1-\alpha$.
This result comes from Proposition~\ref{prop:simple}. However, we are not interested in the sum over $\Scal_{K+1}$. Instead, we want to show that the sum over $\Stest_{K+1}$ has the same property. We need the following assumption: the index is randomly assigned to the calibration set or test set with equal probability:
\begin{align}\label{eq:symmetry}
\mathbb P(i\in \Ical)=\mathbb P(i\in \Itest) = 0.5
\end{align}
for all $i\in\mathcal I\backslash\Itrain$ independently. In other words, $\Ical$ has a uniform distribution on the power set $2^{\mathcal I\backslash\Itrain}$. We summarize the above argument in the following theorem.
\begin{theorem}\label{thm:disjoint}
    Suppose
    \begin{itemize}
        \item[(a)] $S_1, \dots, S_K, S_{K+1}$ are disjoint.
        \item[(b)] The groups are exchangeable in the sense of~\eqref{eq:group:exchange}.
        \item[(c)] The labels in $\Ical\cup\Itest$ are assigned to $\Ical$ and $\Itest$ with equiprobability independently. 
    \end{itemize}
    Then the coverage probability satisfies 
    \begin{align*}
        \mathbb P\left(\sum_{i\in \Stest_{K+1}}y_i\in \mathcal C(\Stest_{K+1})\right)\ge 1-\alpha. 
    \end{align*}
\end{theorem}

\begin{figure}[htbp]
\centering
\begin{tikzpicture}[scale=0.8]
\node[left] at (0,0.5) {\textbf{Proposition 1}};
\draw (0,0) grid (5,1);
\foreach \x in {1,...,4}
{
    \ifnum\x=3
        \node at (\x-0.5,0.5) {$\cdots$};
    \else
        {
        \ifnum\x=4
            \node at (\x-0.5,0.5) {$S_K$};
        \else
            \node at (\x-0.5,0.5) {$S_\x$};
        \fi
        }
    \fi
}
\fill[gray] (4,0) rectangle (5,1); %
\node at (4.5,0.5) {$S_{K+1}$}; %
\draw[decorate,decoration={brace,amplitude=5pt,mirror}] (0,-0.2) -- (3.9,-0.2) node[midway,below,yshift=-5pt] {$\Ical$};
\draw[decorate,decoration={brace,amplitude=5pt,mirror}] (4,-0.2) -- (5,-0.2) node[midway,below,yshift=-5pt] {$\Itest$};

\node[left] at (0,-1.5) {\textbf{Theorem 1}};
\draw (0,-2) grid (5,-1);
\draw (0,-3) grid (5,-2);

\foreach \x in {1,...,4}
{
    \ifnum\x=3
        \node at (\x-0.5,-1.5) {$\cdots$};
    \else
        {
        \ifnum\x=4
            \node at (\x-0.5,-1.5) {$\Scal_K$};
        \else
            \node at (\x-0.5,-1.5) {$\Scal_\x$};
        \fi
        }
    \fi
}
\fill[gray] (4,-2) rectangle (5,-1); %
\node at (4.5,-1.5) {$\Scal_{K+1}$}; %

\foreach \x in {1,...,4}
{
    \ifnum\x=3
        \node at (\x-0.5,-2.5) {$\cdots$};
    \else
        {
        \ifnum\x=4
            \node at (\x-0.5,-2.5) {$\Stest_K$};
        \else
            \node at (\x-0.5,-2.5) {$\Stest_\x$};
        \fi
        }
    \fi
}
\fill[gray] (4,-3) rectangle (5,-2); %
\node at (4.5,-2.5) {$\Stest_{K+1}$}; %

\draw[decorate,decoration={brace,amplitude=5pt}] (5.1,-1) -- (5.1,-2) node[midway,right,xshift=5pt] {$\Ical$};
\draw[decorate,decoration={brace,amplitude=5pt}] (5.1,-2) -- (5.1,-3) node[midway,right,xshift=5pt] {$\Itest$};
\end{tikzpicture}
\caption{Diagram illustrating the partition of data into calibration and test sets for Proposition \ref{prop:simple} and Theorem \ref{thm:disjoint}. The sums within each of $S_1,\dots, S_K, S_{K+1}$ (\textbf{Top}) serve as the calibration data to predict the sum within $S_{K+1}$, mirroring the calibration row for the theorem's diagram (\textbf{Bottom}). As $\Scal_{K+1}$ and $\Stest_{K+1}$ are exchangeable, their sums have the same distribution. Therefore, the prediction set~\eqref{eq:pred:symmetric:test} can be derived similarly to~\eqref{eq:pred:symmetric}.}
\label{fig:data_partition}
\end{figure}
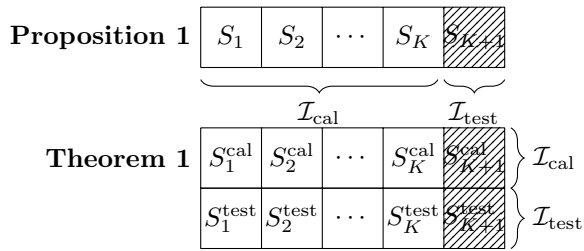

\begin{remark}
    The random splitting assumption~\eqref{eq:symmetry} is slightly different from the procedure of holding out a calibration set $\Ical$ with the same size as $\Itest$ in Algorithm~\ref{alg:cia}. Theorem \ref{thm:disjoint} suggests that it only requires the calibration set and test set have roughly the same size. Suppose $\Ical$ and $\Itest$ are partitions of $\Ical\cup\Itest$ with equal size, this complicates the proof of the theorem, as we can only swap $\Scal_{k}$ and $\Stest_{k}$ when they have the same size.
\end{remark}

\section{Extensions}

In this section, we will extend the applicability of the proposed method to handle overlapping subsets. Furthermore, we will introduce a stratified technique and utilize Conformal Quantile Regression (CQR)~\cite{romano2019conformalized} to enhance the efficiency of our approach.

\subsection{Overlapping Subsets}

Algorithm~\ref{alg:cia} can be applied to overlapping subsets $S_1, \dots, S_K, S_{K+1}$, but the conclusion of Theorem~\ref{thm:disjoint} is valid only if the subsets are disjoint. There are applications where we must assume that the subsets overlap. We will study these applications in Section~\ref{sec:path}. Theoretically, overlapping subsets affect the validity of the method. The following theorem shows that as long as each subset overlaps with only a small portion of the other subsets, the effect on the coverage probability is acceptable.

\begin{theorem}\label{thm:overlap}
    We still assume condition (b) and (c) in Theorem~\ref{thm:disjoint}, and suppose that 
    \begin{align*}
        \max_{\ell\in[K+1]}\frac 1{K+1}\sum_{k\ne\ell} \mathds 1\{S_k\cap S_\ell \ne\emptyset\} = \delta,
    \end{align*}
    Then for fixed $k\in[K+1]$, the coverage probability satisfies 
    \begin{align*}
        \mathbb P\left(\sum_{i\in \Stest_{K+1}}y_i\in \mathcal C(\Stest_{K+1})\right)\ge 1-\alpha-\delta. 
    \end{align*}
\end{theorem}

\subsection{Stratified Conformal Prediction}

Given the known number of unknown labels in each test set, we can leverage this information to enhance the efficiency of conformal prediction. For instance, consider a case where $|\Stest_{k}|$ is relatively small, such as $|\Stest_{k}|=1$. In this case, it is more appropriate to compare the residual with other residuals that have a smaller number of unknown labels in the calibration set. By incorporating this idea, we can modify Algorithm~\ref{alg:cia} and derive a stratified version of the algorithm that takes advantage of this additional information to improve its efficiency.
\begin{algorithm}
\caption{Stratified CIA with Symmetric Calibration}
\begin{algorithmic}[1]
\STATE \textbf{Input:} labels and predictions $\{(y_i, \hat y_i)\}_{i\in \mathcal I\backslash\Itrain}$, index sets $S_1, \dots, S_K, S_{K+1}\subseteq \mathcal I$.  Partition of positive integers $C_1, C_2, \dots$
\STATE \textbf{Output:} Prediction sets for $\sum_{i\in \Stest_{K+1}} y_i$. 
\STATE $\scal_k \leftarrow \left|\sum_{i\in \Scal_k} (y_i-\hat y_i)\right|$ for $k\in[K+1]$.
\STATE Find $C_j$ such that $|\Stest_{k}|\in C_j$.  
\STATE $n_j\leftarrow |\{ k\in[K]: |\Scal_{k}|\in C_j \}|$.
\STATE $Q^j_{1-\alpha}\leftarrow\lceil(1+n_j)(1-\alpha)\rceil$-th smallest value of $\{\scal_k: |\Scal_{k}|\in C_j, k\in[K]\}\cup\{\infty\}$.
\STATE $\mathcal C(\Stest_{K+1})\leftarrow \left[ \sum_{i\in \Stest_{K+1}} \hat y_i-Q^j_{1-\alpha}, \sum_{i\in \Stest_{K+1}} \hat y_i+Q^j_{1-\alpha} \right].$
\STATE \textbf{Output:} $\mathcal C(\Stest_{K+1})$. 
\end{algorithmic}\label{alg:stratified}
\end{algorithm}

In practice, the stratified approach can often reduce the size of the prediction set. However, if the size of $C_j$ is too small, it may negatively impact the validity of the method. To mitigate this issue, it is advisable to set lower bounds for $n_j$ when defining the sets $C_j$. This ensures that each stratum has a sufficient number of samples to maintain the desired coverage probability while still benefiting from the increased efficiency provided by the stratified technique.

\subsection{Conformal Quantile Regression}\label{sec:cqr}

From the theorems, we can observe that the form of the score function can be modified. For instance, we can adopt Conformal Quantile Regression (CQR)~\cite{romano2019conformalized}. The score function can be defined as
\begin{align*}
    \scal_k = \max\left\{ \sum_{i\in \Scal_k} (\hat y_i^{\alpha/2} - y_i), \sum_{i\in \Scal_k} (y_i-\hat y^{1-\alpha/2}_i) \right\}
\end{align*}
for $k\in [K]$, where $\hat y^{\alpha/2}_i$ and $\hat y^{1-\alpha/2}_i$ are the outputs of quantile regression at quantiles $\alpha/2$ and $1-\alpha/2$, respectively, to predict $y_i$. We can obtain $Q_{1-\alpha}$ in Algorithm~\ref{alg:cia} using the score functions of CQR. The prediction set $\mathcal C(\Stest_{k})$ is then given by
\begin{align*}
    \left[ \sum_{i\in \Stest_{k}} \hat y^{\alpha/2}_i-Q_{1-\alpha}, \sum_{i\in \Stest_{k}} \hat y^{1-\alpha/2}_i+Q_{1-\alpha} \right].
\end{align*}
CQR can also be employed in Algorithm~\ref{alg:stratified}, with similar modifications required in Step 4 and Step 7 of the algorithm.

In CQR, if $\hat y_i^{\alpha/2}$ and $\hat y_i^{1-\alpha/2}$ are the true quantiles of the distribution of $Y_i$, then CQR can achieve conditional $1-\alpha$ coverage. However, after combining with CIA, we do not have conditional validity because the sum of quantiles is not necessarily the quantile of the sum. Nevertheless, the upper and lower quantiles can still help in measuring the variation of the estimation and improving the efficiency of the conformal prediction.

\section{Related Work}

In this section, we will explore the relationships and distinctions between the present work and prior research. Although we maintain that the ideas of Conformal Interval Arithmetic (CIA), symmetric calibration, and stratified conformal prediction are original and groundbreaking, and that they are unparalleled among existing techniques for addressing the issues presented in this paper, there are studies in the literature that are relevant to these concepts.

\paragraph{Multiple Test. } Conformal prediction has been applied to multiple testing~\cite{fisch2021efficient, jin2023selection,bates2023testing,bashari2024derandomized,angelopoulos2024conformal}. While our present work also deal with multiple random variables, CIA provides only one confidence interval for the sum of these random variables, so ours is essentially different from these multiple test-based approaches. 

\paragraph{Half Sampling. } In Theorem~\ref{thm:disjoint}, the conclusion can be restated as follows: the $(1-\alpha)$-th quantile of the score function values in the calibration set is larger than a proportion of $1-\alpha$ of the score function values in the test set. At a higher level, this is a comparison between the $(1-\alpha)$-th quantile of half of the samples and the entire sample set. In this regard, symmetric calibration can be linked to the balanced half-sample method~\cite{kish1970balanced,shao1992asymptotic}, which was popular from 1970 to 2000. The balanced half-sample method shares the same idea as symmetric calibration, as it compares the statistic from the half-sample with the entire sample. It is possible that further theory about symmetric calibration can be developed by building upon classical half-sample theory.

\paragraph{Fair Conformal Prediction.} Fair conformal prediction~\cite{lu2022fair, liu2022conformalized, wang2024equal} and group-conditional conformal prediction~\cite{romano2020malice, feldman2021improving, melki2023group, plassier2024conditionally} aim to equalize coverage across various groups characterized by group attributes or auxiliary data. This idea is related to the stratified conformal prediction discussed in this paper. The procedures in these methods are very similar. However, the motivation behind stratified conformal prediction is to improve the efficiency of the conformal prediction, while group-conditional conformal prediction is designed to ensure fairness among different groups.

\paragraph{Robust Route Planning.} As mentioned in the introduction, our method can be applied to path cost prediction. We will further discuss this application in Section~\ref{sec:path}. In~\cite{sun2023predict,patel2024conformal}, the authors introduce Conformal-Predict-Then-Optimize (CPO) for linear programming by predicting edge weights and minimizing upper bounds of path costs for optimal robust solutions. While our method can generate conformal predictions for algorithm-selected paths, choosing the path with the smallest upper bound depends on the score function, compromising the guarantee of $1-\alpha$ coverage. Additionally, applying CPO to our problem is inefficient, as it only provides conformal predictions for the joint distribution of edge weights. We are also aware of recent works \cite{bertsimas2018data, chassein2019algorithms, johnstone2021conformal, stanton2023bayesian, sadana2024survey} which address decision making under uncertainty and methods which tackle adversary data \cite{gendler2021adversarially, bastani2022practical, ghosh2023probabilistically} and distribution shifts \cite{tibshirani2019conformal, prinster2024conformal, zhao2024robust} in conformal prediction.

\section{Application and Experiment}\label{sec:exp}

In this section, we will introduce two main applications of the proposed methods. Then we will analyze the empirical performance of our algorithms on public datasets. 

\subsection{Application 1: Group Average Prediction}\label{sec:group}

Suppose that in a dataset, $x_{ij}$ is the $j$th feature of the $i$th sample, and suppose that this feature is categorical. We can then define the subsets as
\begin{align*}
    S_{k} = \{i\in\Ical\cup\Itest: x_{ij} = k\}
\end{align*}
To predict the average value in class $k$, we can equivalently predict $\sum_{i\in \Stest_k} y_i$, since $\Stest_k$ is the set of indices corresponding to the unknown labels in class $k$.

\subsubsection{Datasets}
\begin{enumerate}
    \item \textit{Bike Sharing \cite{bikeData}:} This dataset is used to investigate the factors influencing bike rental demand. The grouping features are \textsc{season},	\textsc{workingday}, and \textsc{weather}, which have 272 unique value combinations out of 10886 samples. 
    \item \textit{Community Crime \cite{communityData}:} This dataset is used to predict the per capita violent crime rate of a community using its demographic features. The grouping features are \textsc{state} and \textsc{county}, which have 350 unique value combinations out of 1994 samples.
    \item \textit{Medical Expenditure Panel Survey \cite{mepsData}:} This dataset is used to predict the utilization of medical services based on various features. The grouping features are \textsc{age}, \textsc{race}, and \textsc{marriage}, which have 302, 302, and 301 unique value combinations out of 15785, 17541, and 15656 samples for the years 2019, 2020, and 2021, respectively. 
\end{enumerate}
Following similar procedures in \cite{sesia2020comparison}, we standardize the response variables $Y$ for all datasets. We use $70\%$ of data for training the quantile regression model and $30\%$ for calibration and testing.

\subsection{Application 2: Path Cost Prediction}\label{sec:path}

As discussed in the introduction, our method has a valuable application in route planning problems. We consider a transductive setting in edge regression problems, where $\mathcal I$ represents the set of edge indices. In the experiment, we assume that there are edges with unknown labels, and we denote the set of these edges as $\Itest$. Similar to Algorithm~\ref{alg:cia}, we can hold out a set of edges with known labels as the calibration set $\Ical$. Utilizing node and/or edge features, the network structure, and the observed edge labels (costs), we train a Graph Neural Network (GNN) to predict the labels of the calibration set and test set. The transductive setting of GNN is introduced in~\cite{huang2024uncertainty}, where they consider the prediction of node labels. However, more recent works \cite{luo2024conformal, zhao2024conformalized} which focus on edge prediction are more suitable for this particular application set.

While the choice of $S_k$'s can be quite general, the most interesting case in the traffic network data is when we use the edges on the shortest path as the set of edges $S_k$. First, we randomly sample two nodes $(s,t)$ as the source and target nodes, respectively. Then, we use $y_i$ for $i\in \Itrain$ and $\hat y_i$ for $i\in \Ical\cup\Itest$ as the cost of each edge. Next, we find the shortest path using Dijkstra's algorithm. By sampling $K$ pairs of $(s,t)$, the indices of edges on the path form an index set $S_{(s,t)}$. Given the index sets and the predictions $\hat y_i$ for calibration samples and test samples, we can apply Algorithm~\ref{alg:cia} or Algorithm~\ref{alg:stratified} to obtain the conformal prediction.

In this setting, the index sets can overlap, meaning that different paths can contain the same edge. However, in a large network without bottlenecks, two shortest paths with random sources and targets share the same edge with a small probability. Therefore, Theorem~\ref{thm:overlap} can guarantee that the coverage of the prediction set is very close to $1-\alpha$.

\paragraph{Dataset:}  \textit{Road Traffic in Anaheim and Chicago \cite{bar2021transportation}.} This dataset is used to predict the traffic flow along certain roads based on the traffic flow along other roads. The predicted traffic flow is then utilized as edge cost for shortest path planning. This scenario corresponds to the situation of subsets with overlaps, where the edges (roads) in different subsets (shortest paths) may overlap with each other. We collect 2000 shortest paths by randomly sampling $(s, t)$ pairs. The Anaheim dataset consists of 413 nodes and 858 edges, and the Chicago dataset consists of 541 nodes and 2150 edges. We adopt a similar procedure from \cite{jia2020residual, huang2024uncertainty}, and allocate 50\%, 10\%, and 40\% for training, validation, and calibration and testing.

\subsection{Baseline}\label{sec:baseline}

Despite of the significance of our method in application, there are few method can provide valid confidence interval for finding the intervals. We introduce three possible methods for comparison. 

\paragraph{Group Sampling Conformal Prediction. } This method corresponds to the approach described in Section~\ref{sec:cia}. To predict $\sum_{i\in \Stest_{k}} y_i$, we sample $K+1$ groups of samples from the calibration sets, where each group contains $|\Stest_{k}|$ samples. Let $S_1, \dots, S_{K}, S_{K+1}$ be the index sets of these groups. The prediction set is then given by~\eqref{eq:pred:simple}.

It is important to note that in this method, the indices of $S_1, \dots, S_{K}, S_{K+1}$ may not belong to the same class, which means that this approach may not be appropriate in all cases. However, the method is always applicable, regardless of the class composition of the sampled groups.

\paragraph{Normal Confidence Interval.} This approach assumes that $\hat y_i-y_i$ follows an i.i.d. $N(0,\sigma^2)$ distribution. The predicted common variance is estimated as
\begin{align*}
    \hat\sigma^2 = \frac{1}{|\Ical|-1} \sum_{i\in\Ical} (\hat y_i-y_i)^2. 
\end{align*}
Then, the prediction set for $\sum_{i\in \Stest_{k}} y_i$ is given by
\begin{align*}
    \left[ \sum_{i\in \Stest_{k}} \hat y_i+z_{\frac{\alpha}{2}}\sqrt{|\Stest_{k}|}\hat\sigma, \sum_{i\in \Stest_{k}} \hat y_i+z_{1-\frac{\alpha}{2}}\sqrt{|\Stest_{k}|}\hat\sigma\right],
\end{align*}
where $z_{\alpha/2}$ and $z_{1-\alpha/2}$ are the $\alpha/2$ and $(1-\alpha/2)$-th quantile of the standard normal distribution. It is important to note that this approach relies on the assumption of normality, which is unlikely to hold in most real-world datasets.

\paragraph{Bonferroni correction.} The Bonferroni correction has been employed in various conformal methods~\cite{fisch2021efficient, jin2023selection, cleaveland2024conformal}. As mentioned in Section~\ref{sec:naive}, it is also applicable to our problem. While the Bonferroni correction is a valid approach, it often lacks efficiency in the context of our problem.

\subsection{Results}

\begin{figure}[!htb]
\centering
\begin{subfigure}{0.42\textwidth}
\centering
\includegraphics[width=\textwidth]{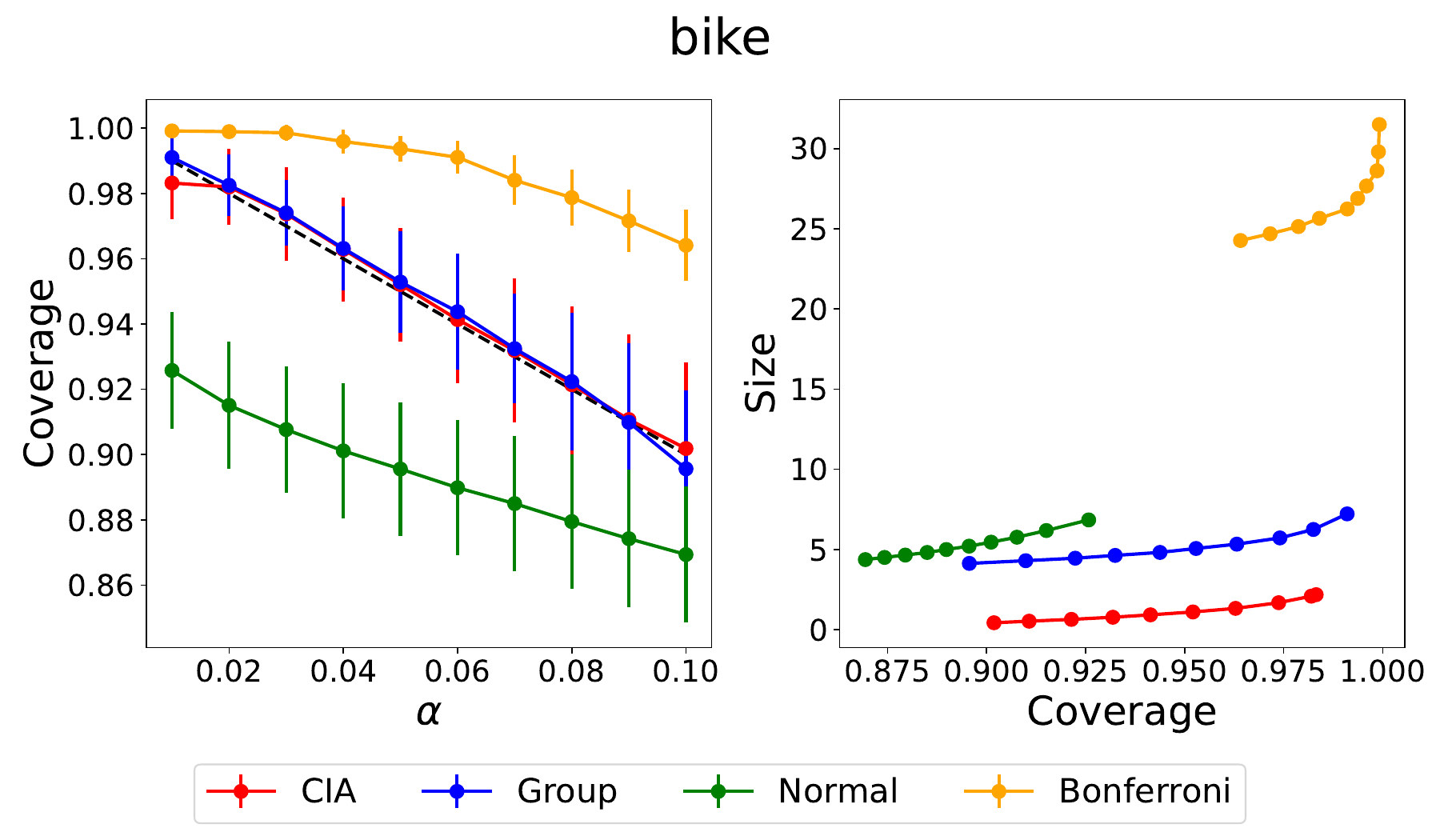}
\end{subfigure}
\hfill
\begin{subfigure}{0.42\textwidth}
\centering
\includegraphics[width=\textwidth]{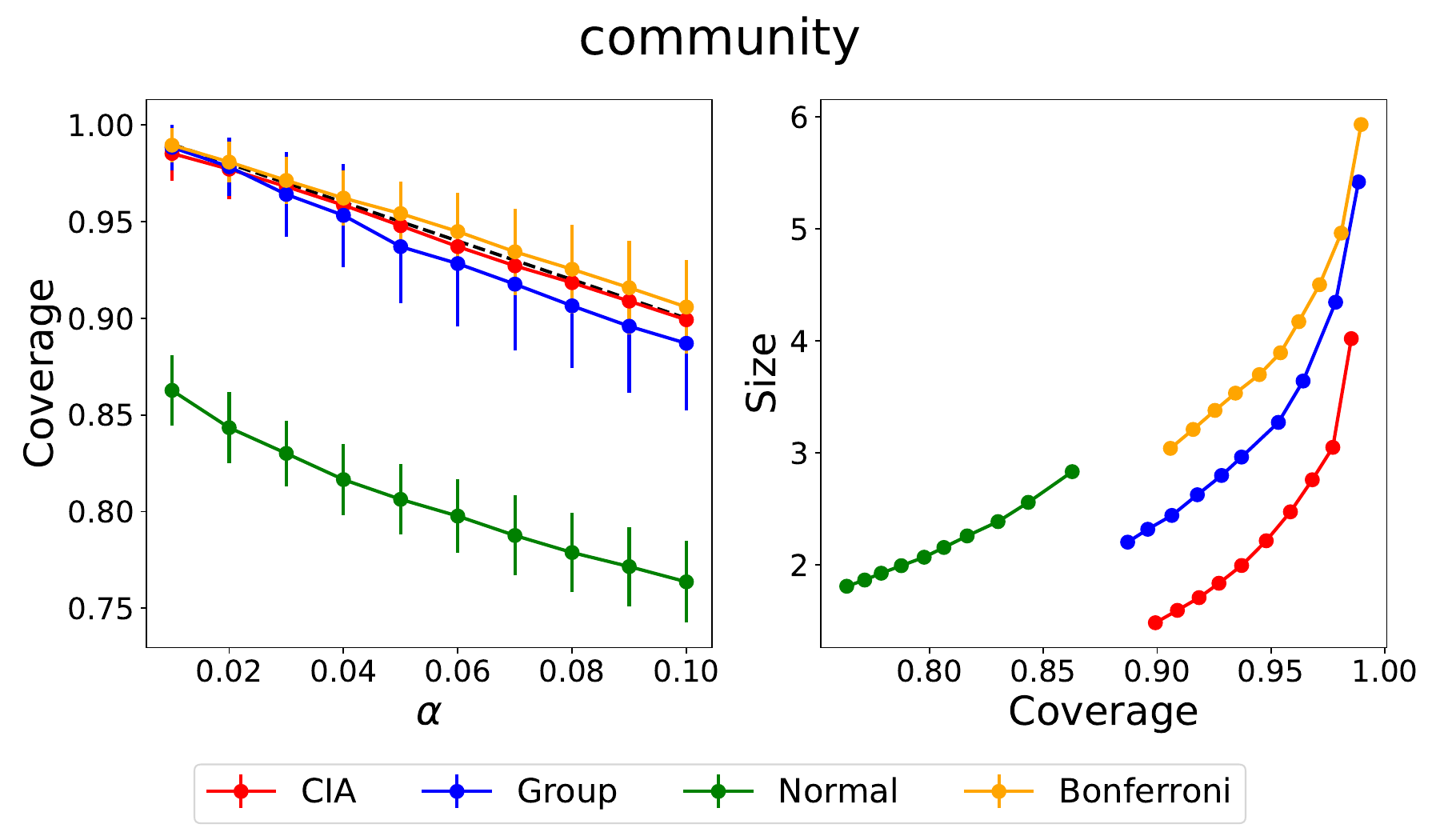}
\end{subfigure}
\hfill
\begin{subfigure}{0.42\textwidth}
\centering
\includegraphics[width=\textwidth]{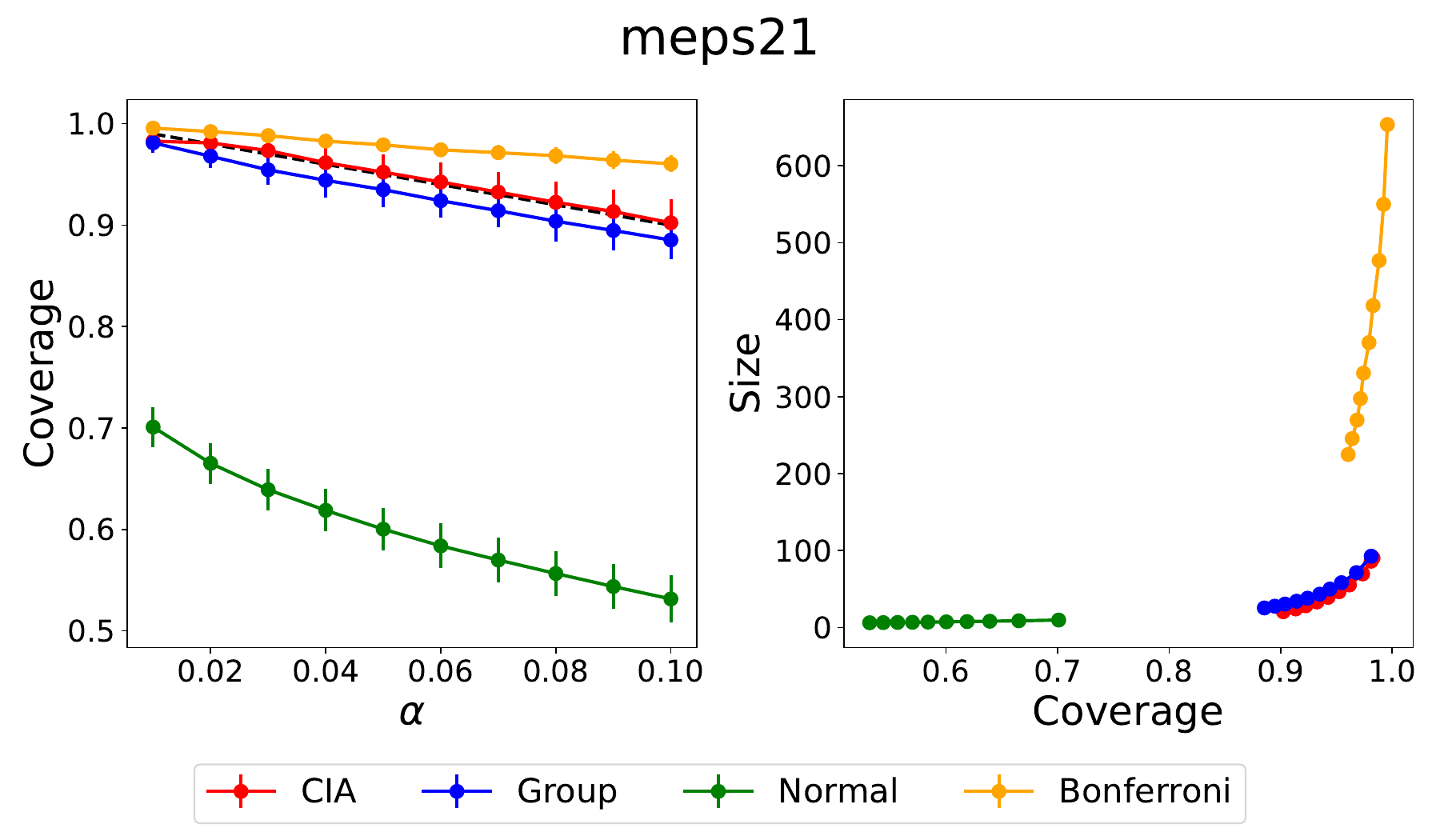}
\end{subfigure}
\caption{Results for constructing prediction sets for subsets without overlaps (Section \ref{sec:group}). CIA achieves guaranteed $1-\alpha$ coverage as well as optimal efficiency on various datasets.}
\label{fig:results_no_overlap}
\end{figure}

We split the calibration and test sets equally as indicated by~\eqref{eq:symmetry}, repeating this process 100 times to record the average results and the standard deviation. Figure \ref{fig:results_no_overlap} presents the results for constructing prediction sets for subsets without overlaps, specifically for the \textit{Bike Sharing}, \textit{Community Crime}, and \textit{Medical Expenditure Panel Survey} datasets. We compare the baseline methods with Algorithm~\ref{alg:stratified} using CQR, as described in Section~\ref{sec:cqr}. In these figures, we refer to our method as ``CIA.'' The baseline methods from Section~\ref{sec:baseline} are labeled as ``Group,'' ``Normal,'' and ``Bonferroni,'' respectively. Additional results will appear in the appendix.

\begin{figure}[!htb]
\centering
\begin{subfigure}{0.42\textwidth}
\centering
\includegraphics[width=\textwidth]{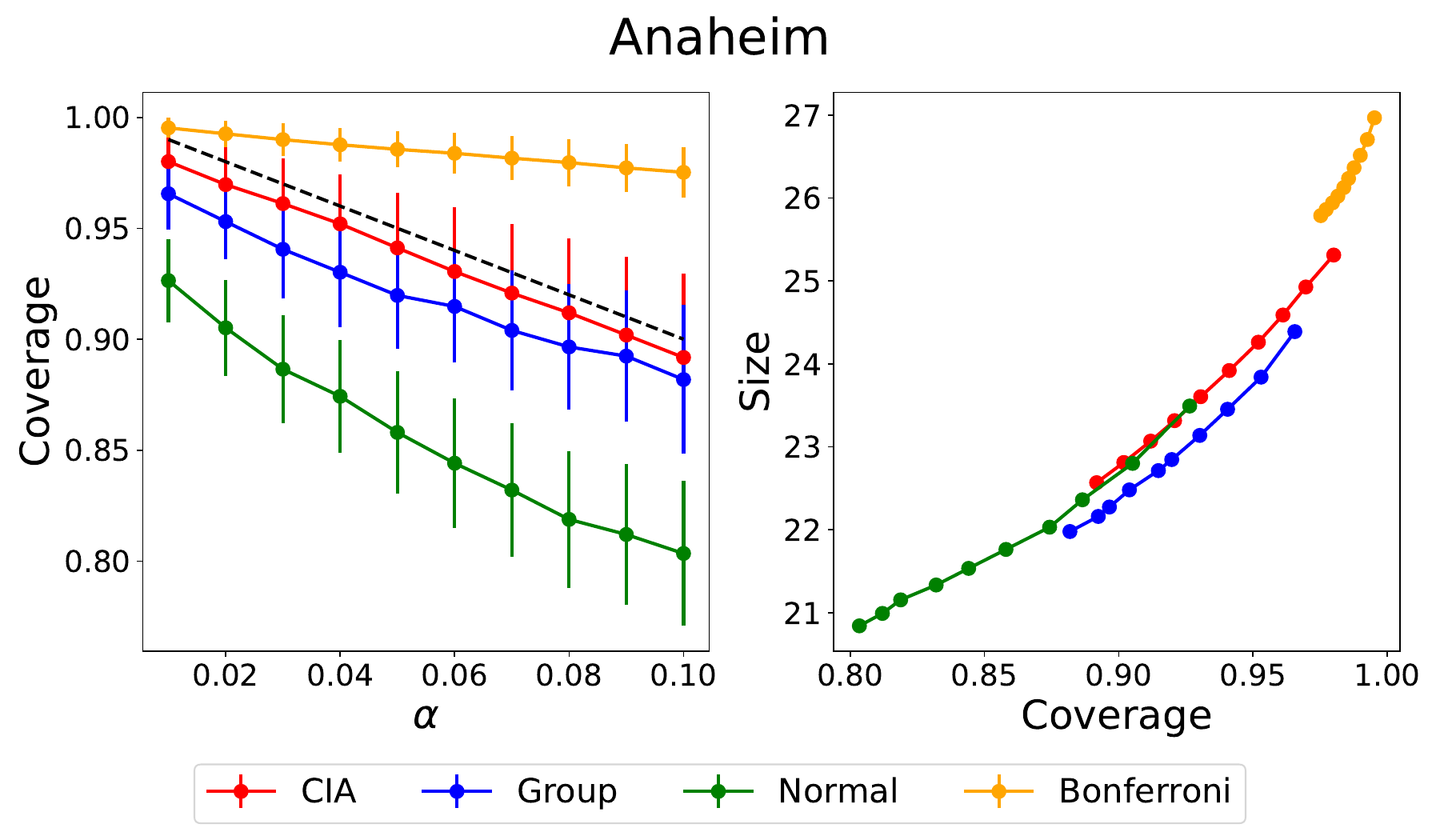}
\end{subfigure}
\hfill
\begin{subfigure}{0.42\textwidth}
\centering
\includegraphics[width=\textwidth]{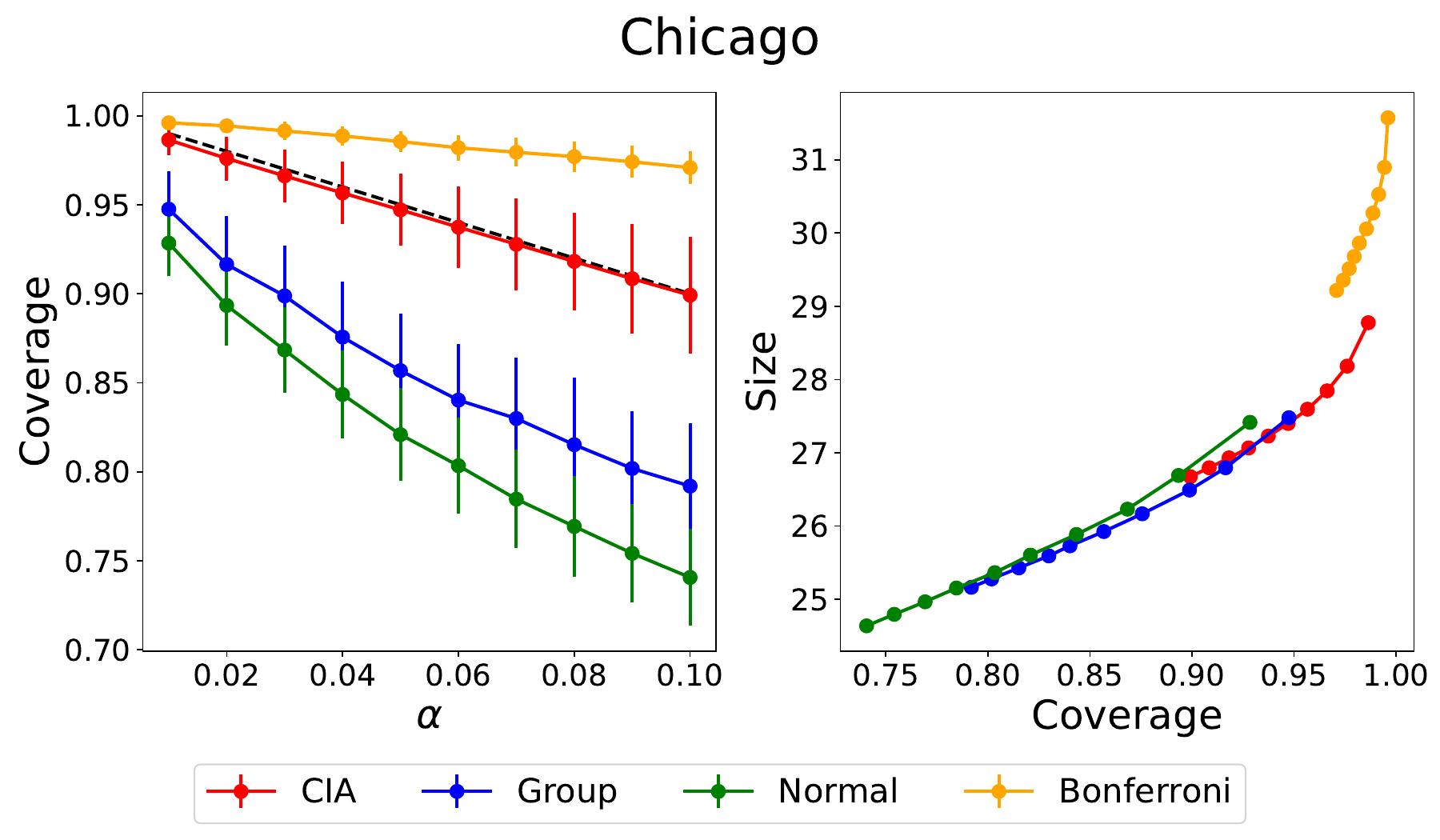}
\end{subfigure}
\caption{Results for constructing prediction sets for subsets with overlaps (Section \ref{sec:path}). CIA achieves close to $1-\alpha$ coverage and optimal efficiency on two road traffic datasets.}
\label{fig:results_overlap}
\end{figure}

For these five datasets, the left plots present the desired miscoverage rate $\alpha$ versus the true coverage rate. The closer the curve aligns with the line $1-\alpha$, the easier it is for the method to achieve the desired coverage. Our proposed method, CIA, is very close to the line $1-\alpha$. There is a small gap in the result of the Anaheim dataset. This result aligns with Theorem~\ref{thm:overlap}. Compared to the Chicago dataset, Anaheim is a smaller traffic network. The probability of two shortest paths overlapping each other can affect the validity. This issue is much less obvious in a larger traffic network like Chicago. The quantity $\delta$ from the theorem, along with the coverage gap, is provided in the appendix. Recall that the group-sample conformal method is proposed in Section~\ref{sec:cia}. This is also a conformal prediction method, so its validity is also close to the desired line $1-\alpha$ in the results of Figure~\ref{fig:results_no_overlap}. However, the validity is much less reliable in the results of Figure~\ref{fig:results_overlap}. This is because edge weights from random samples are not exchangeable with the edge weights on the shortest path. The method of Bonferroni correction has good validity in the example of the community dataset. The reason is that under the current setting of calibration and test set ratio, each subset contains one or two samples, which is very close to the case of classical conformal prediction. In other datasets, Bonferroni correction has too high coverage. For normal confidence intervals, since the assumption is not true in general, the coverage is much lower than the desired one. In conclusion, our proposed method has the most reliable validity in all datasets that we have studied.

Under the choices of $\alpha$ on the left plots, the right plots compare the coverage versus the prediction set size. The lower the curve, the more efficient and informative the prediction set provided by the method. In the plots of Figure~\ref{fig:results_no_overlap}, CIA is the most efficient in all datasets. In the plots of Figure~\ref{fig:results_overlap}, CIA has similar efficiency to the Group and Normal methods. In conclusion, CIA is the most efficient method with valid coverage.

\section{Conclusion}

In this paper, we have introduced conformalized interval arithmetic with symmetric calibration, to tackle the challenges of conformal prediction for interdependent data. Our approach offers methodological contributions and theoretical guarantees for achieving the desired coverage under reasonable exchangeability assumptions. Extensive experiments on real-world datasets demonstrate the validity and efficiency of our method, outperforming others even when they fail to ensure valid coverage. We hope our core ideas and techniques will foster the development of more robust conformal prediction methods across various domains.

\section{Acknowledgments}

The work described in this paper was partially supported by a grant from City University of Hong Kong (Project No.9610639). We thank the reviewers for their insightful suggestions. We also appreciate the helpful discussions with Andro Sabashvili about Theorem~\ref{thm:disjoint}.

\appendix

\section{Proof of Theorem 1}

We first consider a fixed permutation and a fixed partition of calibration and test set. Let $\rcal_{k}$ be the rank of $\scal_{k}$ in $\scal_{k}$ for $k\in[K+1]$.  We exchange the calibration set and test set in $S_{k}$, i.e., we swap $\Stest_{K+1}$ and $\Scal_{k}$. Under the symmetry assumption on the calibration indices and test indices, this swapping procedure corresponds two possible partitions with the same probability. Let $\rtest_{K+1}$ be the rank of $\stest_{K+1}$ in $\{\scal_{k}: k=1, \dots, K\}\cup\{\stest_{K+1}\}$ after swapping. Under the assumption of Theorem 1, the scores of the calibration set remains the same. Hence $\rtest_{K+1}=\rcal_{K+1}$. $\rtest_{K+1}$ has the same distribution as $\rcal_{K+1}$. Hence,
    \begin{align*}
        \mathbb P(\stest_{K+1}\le Q_{1-\alpha})&=\mathbb P(\rtest_{K+1}\le \lceil(1+K)(1-\alpha)\rceil)\\
        &=\mathbb P(\rcal_{K+1}\le \lceil(1+K)(1-\alpha)\rceil)\\
        &\ge 1-\alpha. 
    \end{align*}
    The event $\rtest_{K+1}\le \lceil(1+K)(1-\alpha)\rceil$ is equivalent to $\sum_{i\in\Stest_{K+1}} y_i\in \mathcal C(\Stest_{K+1})$ using the definition of $Q_{1-\alpha}$. This completes the proof.  

\section{Proof of Theorem 2}

The assumption of the theorem means there are most $(K+1)\delta$ many subsets overlaps with $S_{K+1}$. When swapping the $\Scal_{K+1}$ and $\Stest_{K+1}$, it affects at most $(K+1)\delta$ many scores in $\scal_k, k\in[K]$. Thus, $\rtest_{K+1}\le \rcal_{K+1}+(K+1)\delta$. 
    \begin{align*}
        &\mathbb P(\stest_{K+1}\le Q_{1-\alpha})\\
        &= \mathbb P(\rtest_{K+1}\le \lceil (1+K)(1-\alpha)\rceil)\\
        &\ge \mathbb P(\rcal_{K+1}+(K+1)\delta\le \lceil (1+K)(1-\alpha)\rceil)\\
        &=\mathbb P(\rcal_{K+1}\le \lceil (1+K)(1-\alpha-\delta)\rceil)\\
        &\ge 1-\alpha-\delta. 
    \end{align*}
    The proof is complete. 

\section{Summary of Methods and Results}
Due to space limitations, we present one variant of the proposed CIA method using CQR to compute scores, along with three representative baseline methods in the main text. Here, we provide a comprehensive summary of the various CIA method variants and additional baseline methods for comparison.

\subsection{Methods}
\subsubsection{CIA Variants}
\begin{enumerate}[left=0pt, labelsep=10pt]
    \item \textbf{CIA (Split) Nonstratified.} 
    This variant of CIA uses a split conformal prediction approach without stratification. The score function is 
    \begin{align}\label{eq:Split}
    \scal_k := \left| \sum_{i\in \Scal_k} (y_i-\hat y_i) \right|, k\in [K].
    \end{align}

    \item \textbf{CIA (CQR) Nonstratified.} 
    This variant of CIA employs conformal quantile regression (CQR) without stratification. The score function is
    \begin{align}\label{eq:CQR}
    \scal_k = \max\left\{ \sum_{i\in \Scal_k} (\hat y_i^{\alpha/2} - y_i), \sum_{i\in \Scal_k} (y_i-\hat y^{1-\alpha/2}_i) \right\}, k\in [K].
    \end{align}

    \item \textbf{CIA (Split) Stratified.} 
    This variant of CIA utilizes the same score function (\ref{eq:Split}) as the nonstratified version. It stratifies the calibration set based on the number of samples in each group $|\Scal_k|$ when computing the quantile of the scores to determine the prediction set threshold.

    \item \textbf{CIA (CQR) Stratified.} 
    This CIA variant is a stratified version which uses the same score function as in (\ref{eq:CQR}).
\end{enumerate}

\subsubsection{Baseline Using Conformal Prediction}
\begin{enumerate}[left=0pt, labelsep=10pt]
    \item \textbf{Group Sampling Conformal Prediction (Split).} 
    This baseline method uses conformal prediction, sampling groups of size $|\Stest_{K+1}|$ without replacement, as described in Section 7.3 in the main text. It employs the same score function as (\ref{eq:Split}). 

    \item \textbf{Group Sampling Conformal Prediction (CQR).} 
    This baseline method combines the CQR score function (\ref{eq:CQR}) with group  sampling. 

    \item \textbf{Bonferroni Correction (Split).} 
    This baseline method applies Bonferroni correction to the split conformal prediction intervals with the same score function (\ref{eq:Split}). The corrected intervals account for varying sizes of $|\Stest_k|$.

    \item \textbf{Bonferroni Correction (CQR).} 
    This baseline method applies Bonferroni correction to the CQR intervals computed with score function (\ref{eq:CQR}).
\end{enumerate}

\subsubsection{Baseline Using Normal Approximation}
\begin{enumerate}[left=0pt, labelsep=10pt]
    \item \textbf{Normal Confidence Interval with Homoscedasticity Assumption.} 
    This baseline method, introduced in Section 7.3 in the main text, constructs confidence intervals assuming an i.i.d. normal distribution $N(0,\sigma^2)$ for all $\hat y_i-y_i$. The predicted common variance is estimated as
    \begin{align*}
    \hat\sigma^2 = \frac{1}{|\Ical|-1} \sum_{i\in\Ical} (\hat y_i-y_i)^2. 
    \end{align*}

    \item \textbf{Normal Confidence Interval Using IQR.}  
    This baseline method constructs confidence intervals assuming independent zero-mean normal distributions with varying standard deviations for $\hat{y}_i - y_i$. The standard deviation is estimated using the interquartile range (IQR) for each sample, predicted via a quantile regression model:
    \begin{align*}
        \hat\sigma_i = \frac{\hat{q}_{0.75}(x_i) - \hat{q}_{0.25}(x_i)}{z_{0.75} - z_{0.25}},
    \end{align*}
    where $z_{\beta}$ is the $\beta$ quantile of the standard normal distribution. This formulation enables heteroscedasticity.
\end{enumerate}

\subsection{Code Implementation}
The attached code contains the implementation of CIA and other baseline methods. The main file, \texttt{main.py}, runs experiments on the bike, community, and MEPS datasets, while \texttt{plot\_results.py} generates plots of the results. The code relies on dependencies such as Python 3.x, PyTorch, NumPy, Pandas, Matplotlib, and Scikit-learn.

The Split Conformal Method uses a neural network with three fully connected layers, and Conformal Quantile Regression employs quantile regression forests for predictions. For experiments with subsets that have overlaps, the \texttt{gnn} folder contains additional code, including a directed graph autoencoder implemented in \texttt{digae\_traffic.py} and \texttt{autoencoder.py}. Required packages include networkx, geopandas, and torch-geometric.

\subsection{Results}
We present the comparison results of the aforementioned methods across various datasets discussed in the two applications in the experiment section of the main paper. The results demonstrate the performance of each method in terms of coverage probability and set size.

\section{Analysis of CIA on Overlapping Subsets}
In the main text, we have demonstrated the extension of the proposed CIA method to overlapping subsets with an acceptable compromise on the coverage. In particular, Theorem 2 shows that as long as each subset overlaps with only a small portion of the other subsets, the effect on the coverage probability is acceptable.

Here, we conduct an extensive analysis to further investigate the performance of CIA on overlapping subsets. We vary the degree of overlap between subsets and evaluate the impact on the coverage probability and interval width.

For both Anaheim and Chicago datasets, the shortest paths represent the subsets of edges corresponding to individual road segments. The overlap ratio of the subsets is computed as follows:
\begin{align*}
\delta = \frac{1}{\binom{K+1}{2}} \sum_{k \neq \ell, k,\ell \in [K+1]} \frac{|S_k \cap S_\ell|}{|S_k \cup S_\ell|},
\end{align*}
which serves as an averaged version of the \(\delta\) defined in Theorem 2 of the main text for easier numerical analysis. The term \(\frac{|S_k \cap S_\ell|}{|S_k \cup S_\ell|}\) represents the overlap ratio between subsets \(S_k\) and \(S_\ell\), calculated as the size of their intersection divided by the size of their union.

\begin{figure}
    \centering
    \includegraphics[width=0.95\linewidth]{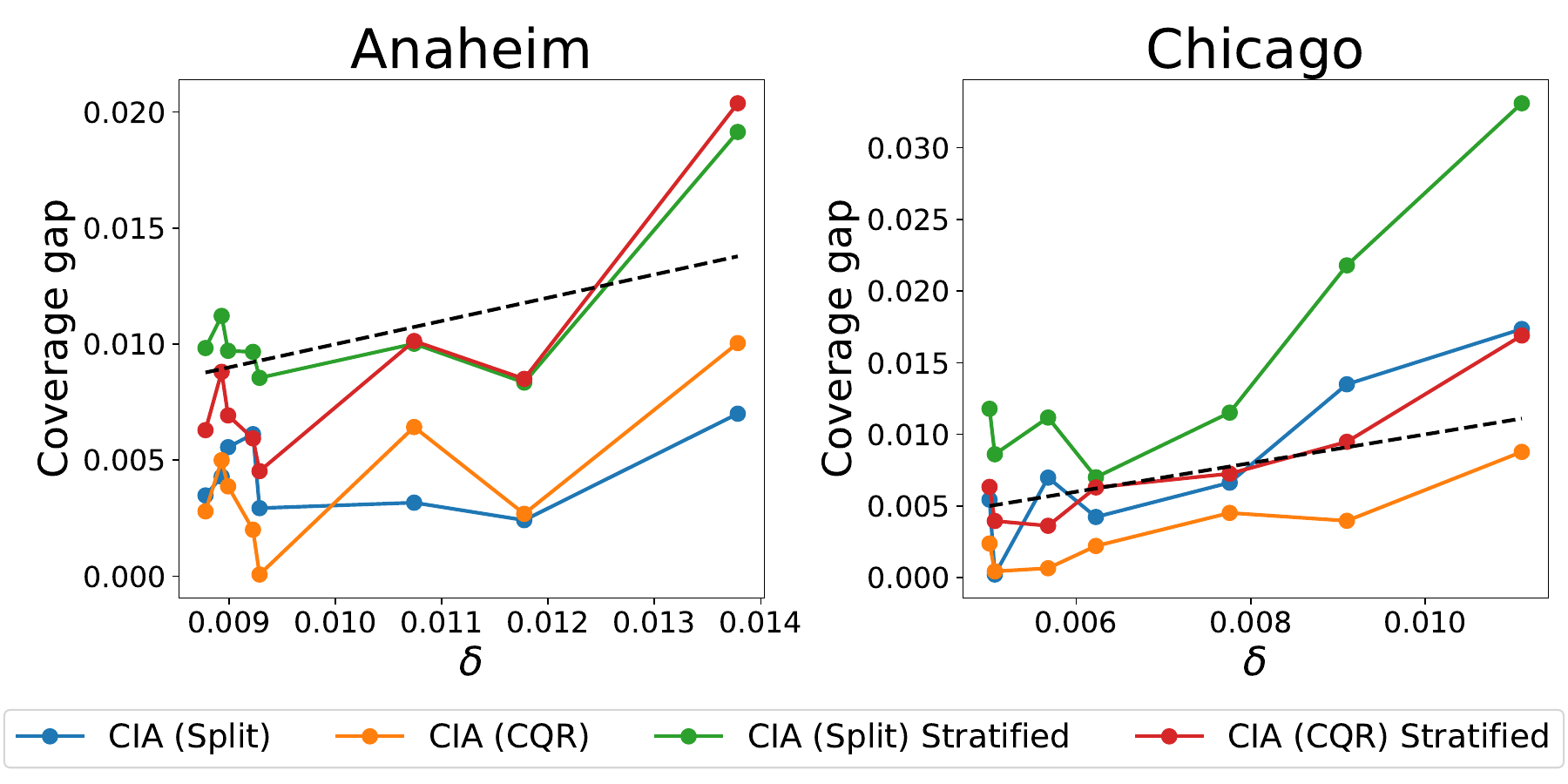}
    \caption{The figure illustrates the positive correlation between the subsets overlap ratio $\delta$ and the coverage gap (the difference between the coverage probability and the desired $1-\alpha$ coverage). }
    \label{fig:coverage gap}
\end{figure}

We alternate the subsets overlap ratio by setting varying lower bounds on the length of the shortest paths. With increasing lower bounds on the length of the shortest paths, there tends to be an increasing overlap between different shortest paths.
In Figure \ref{fig:coverage gap}, we show there is a positive correlation between the subsets overlap ratio $\delta$ and the coverage gap, i.e., the difference between the coverage probability and the desired $1-\alpha$ coverage.
We demonstrate the results for all four variants of the proposed CIA method.

The analysis reveals that CIA maintains good coverage probability even when there is moderate overlap between subsets. The interval width remains relatively stable across different overlap scenarios, indicating the robustness of CIA in handling overlapping subsets.

Overall, the experimental results support the theoretical findings and demonstrate the effectiveness of CIA in providing valid confidence intervals for overlapping subsets, with a reasonable trade-off between coverage probability and interval width.

\begin{figure*}
\centering
\begin{multicols}{2}
    \begin{subfigure}{0.96\linewidth}
    \centering
    \includegraphics[width=\linewidth]{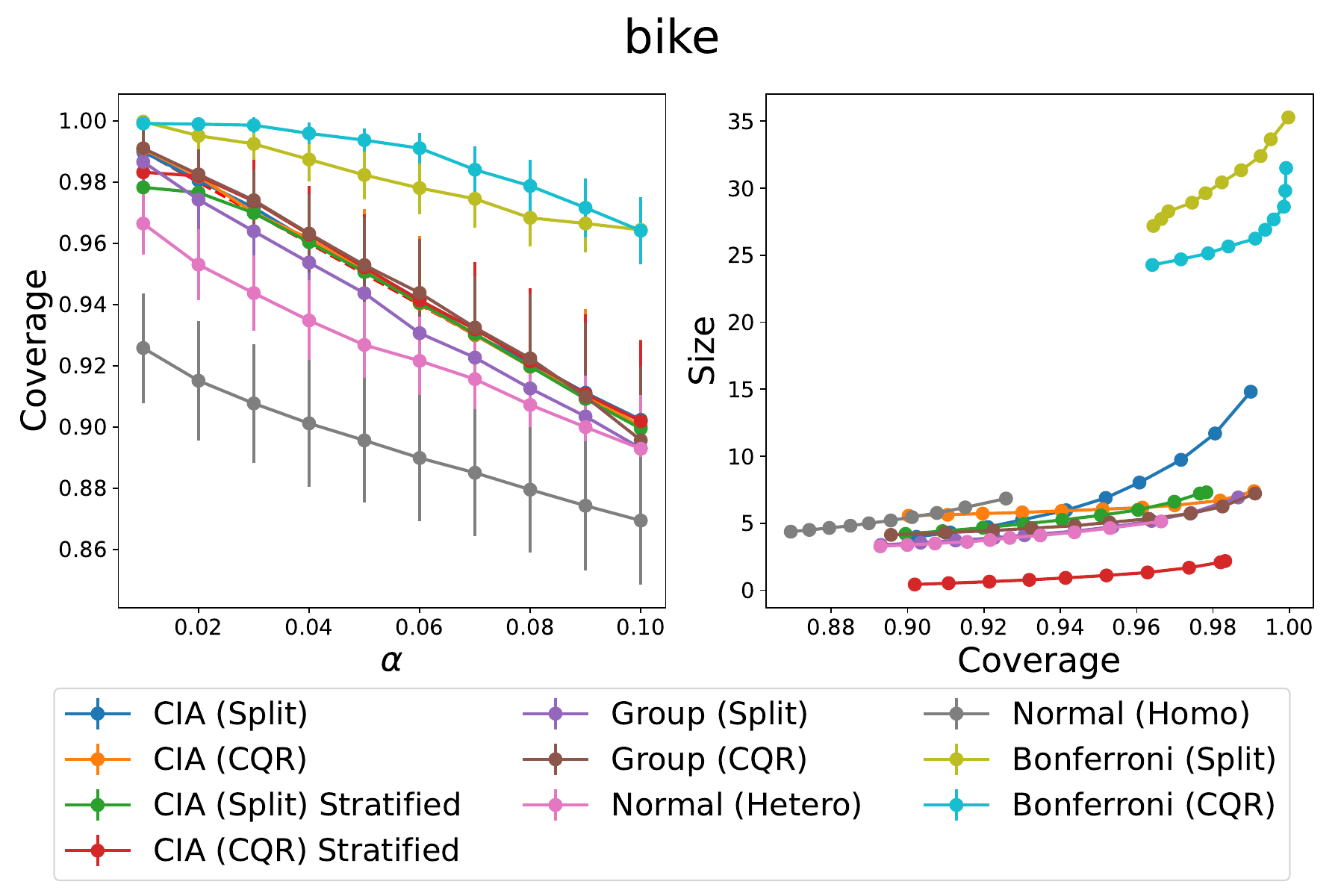}
    \end{subfigure}

    \begin{subfigure}{0.96\linewidth}
    \centering
    \includegraphics[width=\linewidth]{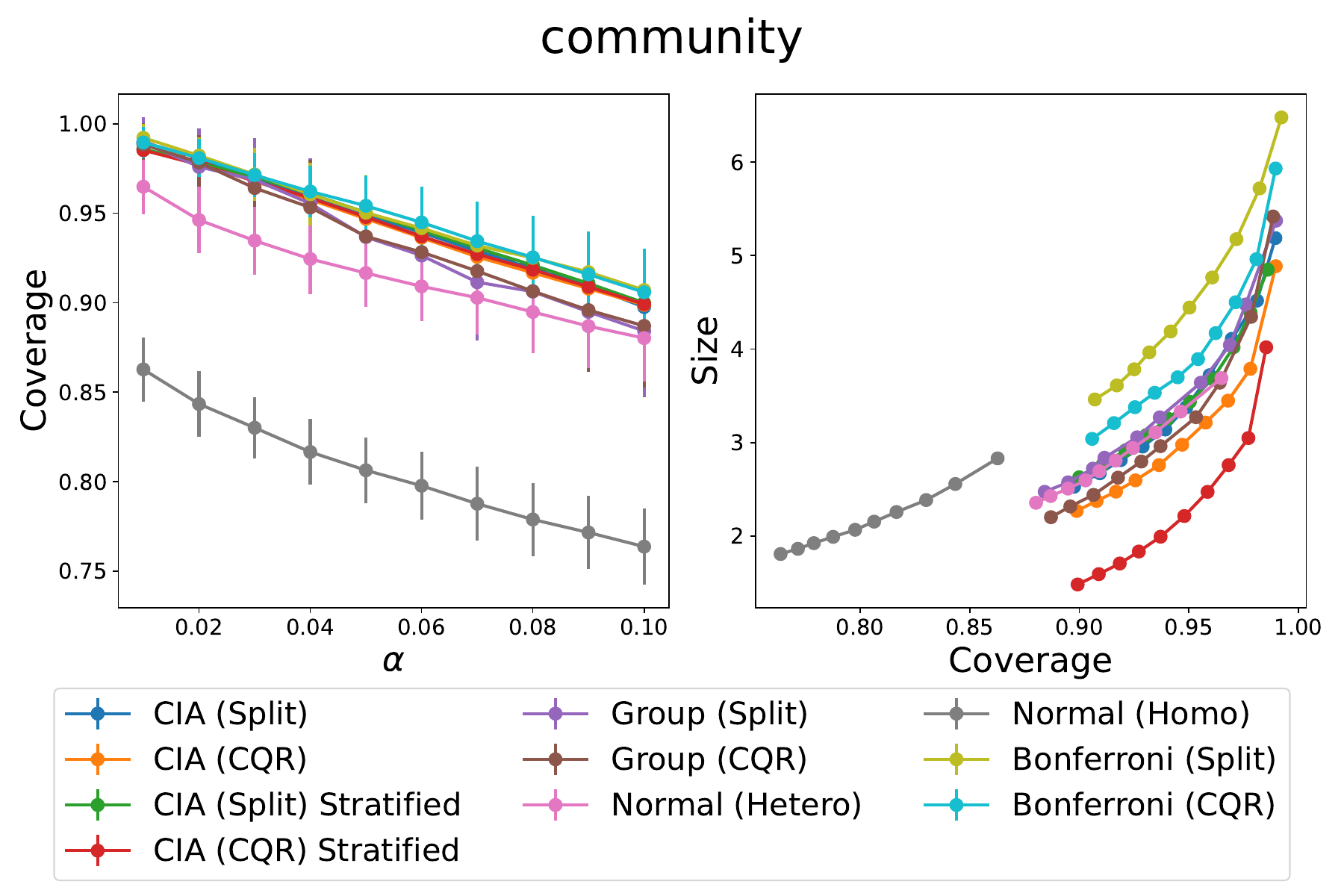}
    \end{subfigure}

    \begin{subfigure}{0.96\linewidth}
    \centering
    \includegraphics[width=\linewidth]{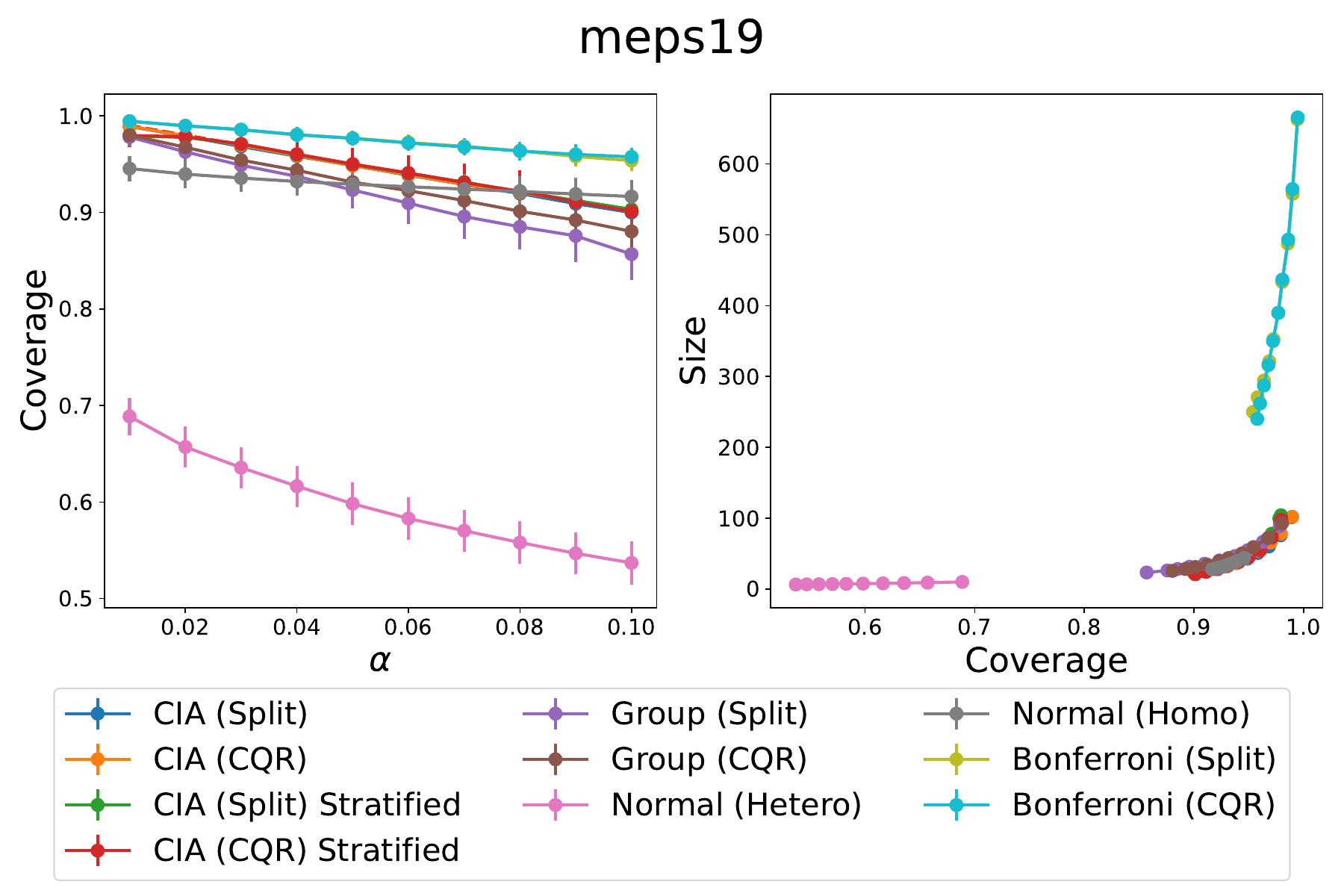}
    \end{subfigure}

    \begin{subfigure}{0.96\linewidth}
    \centering
    \includegraphics[width=\linewidth]{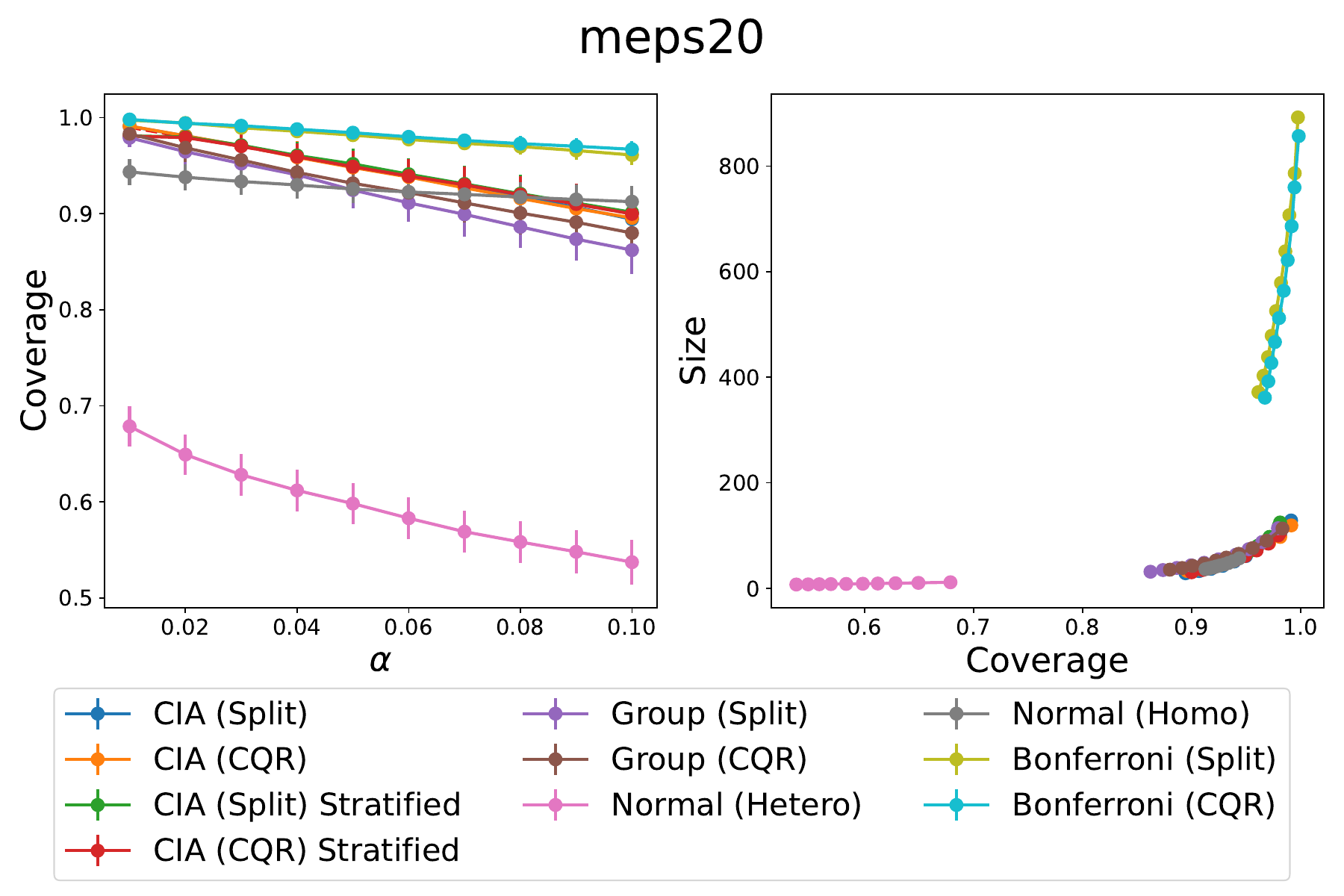}
    \end{subfigure}

    \begin{subfigure}{0.96\linewidth}
    \centering
    \includegraphics[width=\linewidth]{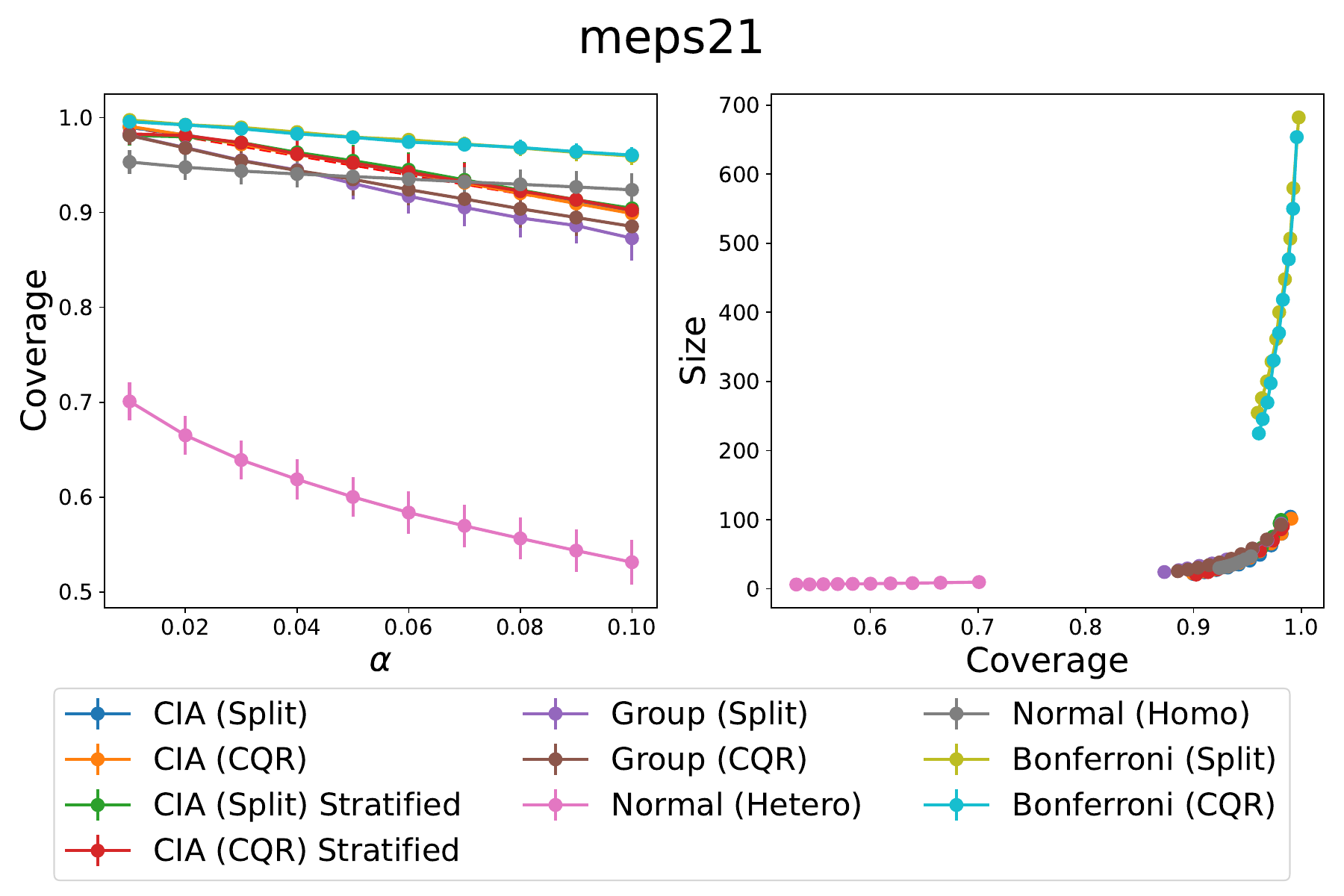}
    \end{subfigure}

    \begin{subfigure}{0.96\linewidth}
    \centering
    \includegraphics[width=\linewidth]{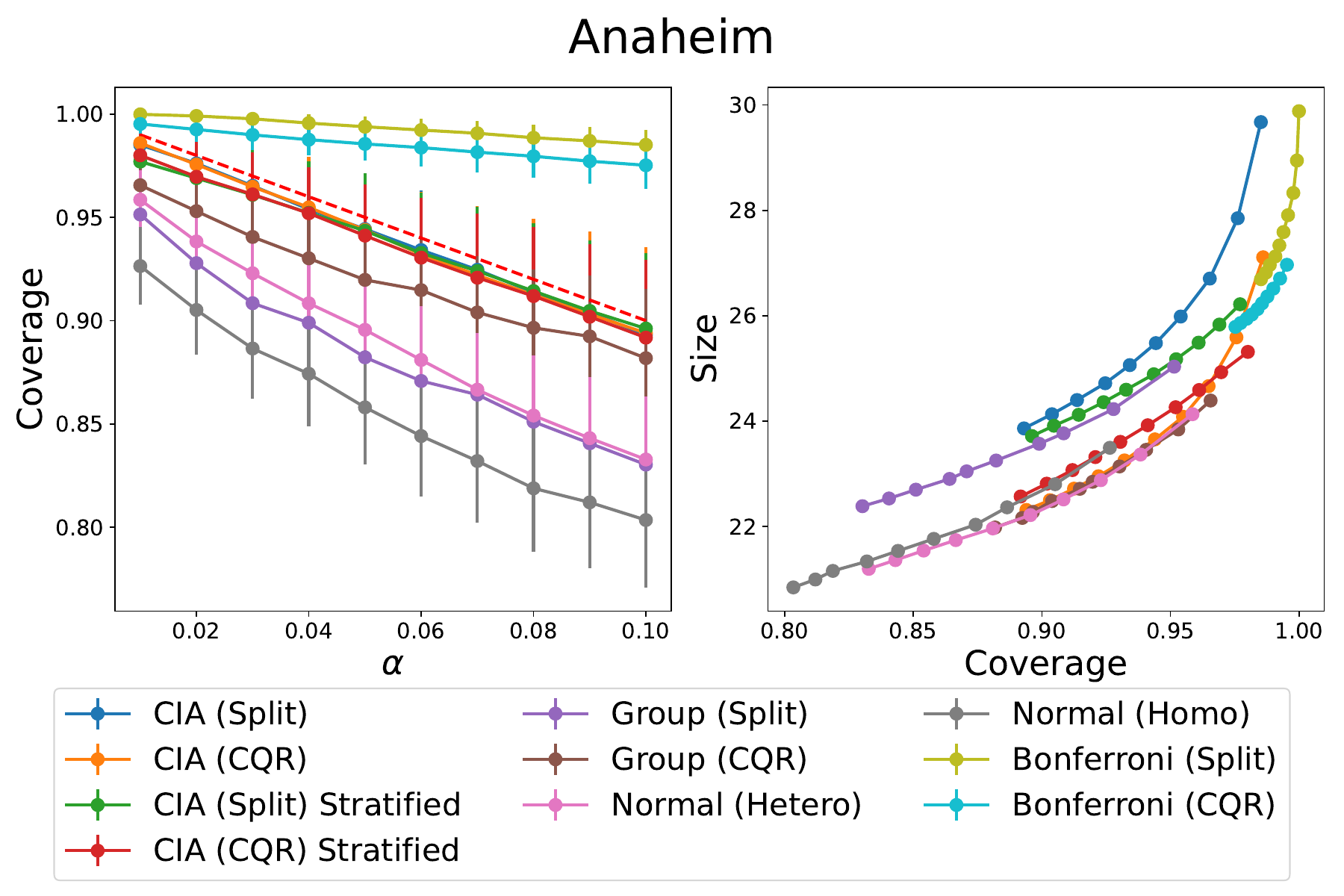}
    \end{subfigure}

    \begin{subfigure}{0.96\linewidth}
    \centering
    \includegraphics[width=\linewidth]{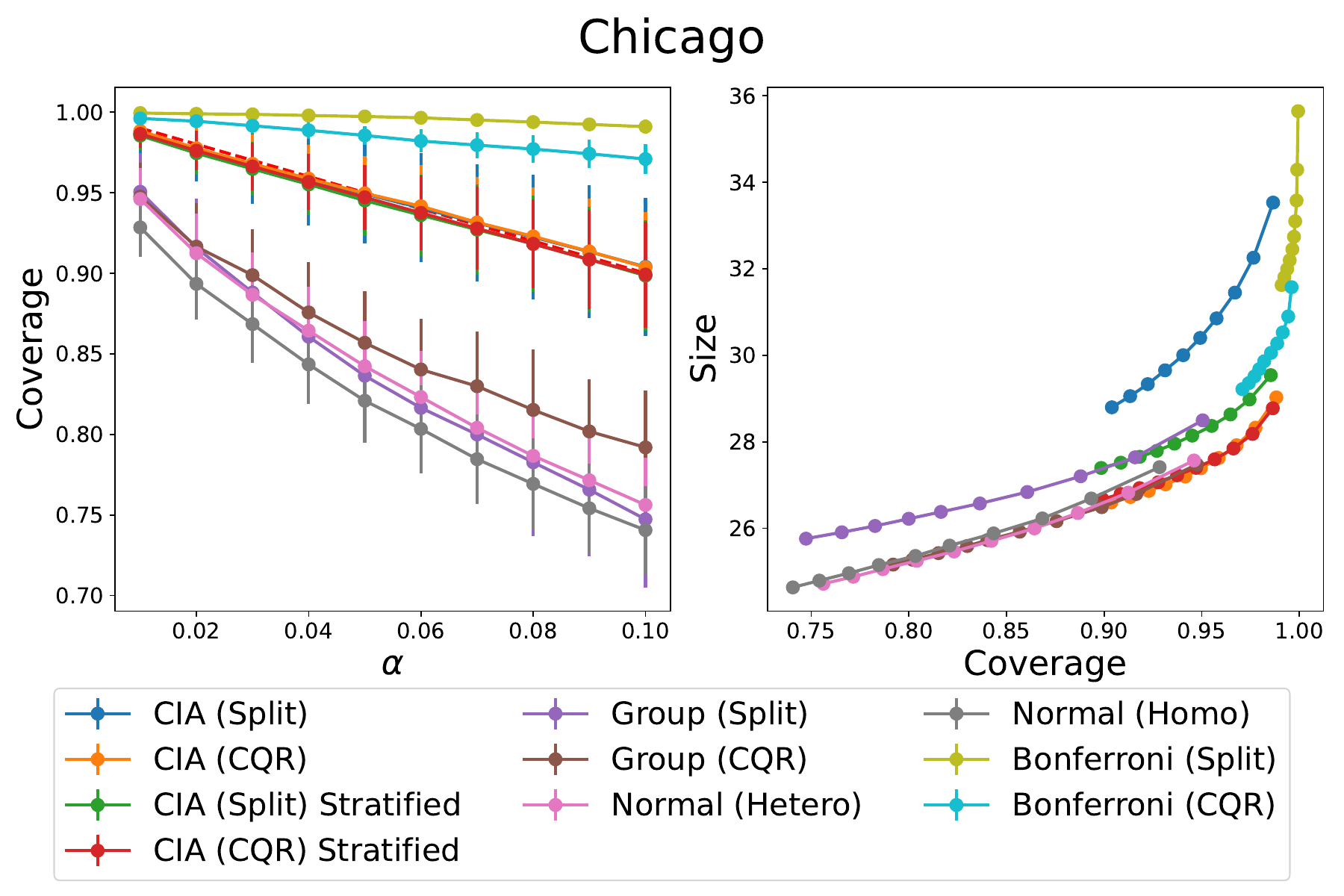}
    \end{subfigure}
\end{multicols}
\caption{Complete results for prediction sets. All CIA variants achieve $1-\alpha$ coverage, with \textbf{CIA (CQR) Stratified} showing optimal efficiency.}
\label{fig:results_combined}
\end{figure*}

\begin{table*}
\centering
\caption{Coverage and Size Results for Different Methods Across Datasets at $\alpha=0.1$}
\begin{adjustbox}{max width=\textwidth}
\begin{tabular}{lcccccccccc}
\toprule
& \multicolumn{2}{c}{\textbf{CIA (Split)}} & \multicolumn{2}{c}{\textbf{CIA (CQR)}} & \multicolumn{2}{c}{\textbf{CIA (Split) Stratified}} & \multicolumn{2}{c}{\textbf{CIA (CQR) Stratified}} & \multicolumn{2}{c}{\textbf{Group (Split)}} \\
\cmidrule(lr){2-3} \cmidrule(lr){4-5} \cmidrule(lr){6-7} \cmidrule(lr){8-9} \cmidrule(lr){10-11}
Dataset & Coverage & Size & Coverage & Size & Coverage & Size & Coverage & Size & Coverage & Size \\
\midrule
Anaheim & 0.89 (0.04) & 23.86 (0.78) & 0.89 (0.04) & 22.31 (1.04) & 0.90 (0.04) & 23.71 (0.97) & 0.89 (0.04) & 22.57 (1.16) & 0.83 (0.04) & 22.38 (0.90) \\
bike & 0.90 (0.02) & 3.89 (0.41) & 0.90 (0.03) & 5.56 (0.16) & 0.90 (0.03) & 4.00 (0.40) & 0.90 (0.03) & 0.42 (0.19) & 0.89 (0.02) & 3.28 (0.18) \\
Chicago & 0.90 (0.04) & 28.80 (1.04) & 0.90 (0.03) & 26.60 (1.49) & 0.90 (0.03) & 27.40 (1.47) & 0.90 (0.03) & 26.67 (1.55) & 0.75 (0.04) & 25.76 (1.35) \\
community & 0.90 (0.03) & 2.53 (0.18) & 0.90 (0.03) & 2.27 (0.17) & 0.90 (0.02) & 2.63 (0.19) & 0.90 (0.03) & 1.48 (0.20) & 0.88 (0.04) & 2.47 (0.30) \\
meps\_19 & 0.90 (0.02) & 22.87 (2.61) & 0.90 (0.02) & 22.76 (2.94) & 0.90 (0.02) & 25.83 (3.35) & 0.90 (0.03) & 20.83 (4.00) & 0.86 (0.03) & 23.33 (2.90) \\
meps\_20 & 0.89 (0.02) & 28.75 (3.45) & 0.90 (0.02) & 33.11 (3.51) & 0.90 (0.02) & 33.83 (4.88) & 0.90 (0.02) & 30.56 (4.78) & 0.86 (0.03) & 31.61 (4.04) \\
meps\_21 & 0.90 (0.02) & 21.39 (2.57) & 0.90 (0.02) & 22.61 (2.82) & 0.90 (0.02) & 25.06 (3.18) & 0.90 (0.02) & 20.50 (3.48) & 0.87 (0.02) & 24.19 (3.22) \\
\midrule
& \multicolumn{2}{c}{\textbf{Group (CQR)}} & \multicolumn{2}{c}{\textbf{Normal (Hetero)}} & \multicolumn{2}{c}{\textbf{Normal (Homo)}} & \multicolumn{2}{c}{\textbf{Bonferroni (Split)}} & \multicolumn{2}{c}{\textbf{Bonferroni (CQR)}} \\
\cmidrule(lr){2-3} \cmidrule(lr){4-5} \cmidrule(lr){6-7} \cmidrule(lr){8-9} \cmidrule(lr){10-11}
Dataset & Coverage & Size & Coverage & Size & Coverage & Size & Coverage & Size & Coverage & Size \\
\midrule
Anaheim & 0.88 (0.03) & 21.98 (0.91) & 0.80 (0.03) & 20.84 (0.82) & 0.83 (0.03) & 21.19 (0.81) & 0.99 (0.01) & 26.69 (1.17) & 0.98 (0.01) & 25.79 (1.13) \\
bike & 0.90 (0.02) & 4.13 (0.19) & 0.87 (0.02) & 4.37 (0.09) & 0.89 (0.02) & 3.18 (0.09) & 0.96 (0.01) & 25.84 (1.06) & 0.96 (0.01) & 24.27 (0.94) \\
Chicago & 0.79 (0.04) & 25.16 (1.35) & 0.74 (0.03) & 24.64 (1.31) & 0.76 (0.03) & 24.72 (1.31) & 0.99 (0.00) & 31.62 (1.95) & 0.97 (0.01) & 29.22 (1.78) \\
community & 0.89 (0.03) & 2.20 (0.27) & 0.76 (0.02) & 1.81 (0.03) & 0.88 (0.02) & 2.36 (0.12) & 0.91 (0.02) & 3.46 (0.22) & 0.91 (0.02) & 3.04 (0.25) \\
meps19 & 0.88 (0.02) & 25.67 (2.84) & 0.54 (0.02) & 6.42 (0.08) & 0.92 (0.02) & 28.01 (2.32) & 0.95 (0.01) & 250.03 (30.56) & 0.96 (0.01) & 240.15 (31.88) \\
meps20 & 0.88 (0.02) & 35.55 (3.67) & 0.54 (0.02) & 7.44 (0.08) & 0.91 (0.02) & 36.45 (2.59) & 0.96 (0.01) & 371.84 (45.07) & 0.97 (0.01) & 361.34 (44.02) \\
meps21 & 0.89 (0.02) & 25.40 (2.67) & 0.53 (0.02) & 6.15 (0.06) & 0.92 (0.02) & 29.99 (2.44) & 0.96 (0.01) & 254.63 (31.54) & 0.96 (0.01) & 224.75 (31.47) \\
\bottomrule
\end{tabular}
\end{adjustbox}
\end{table*}

\begin{table*}
\centering
\caption{Coverage and Size Results for Different Methods Across Datasets at $\alpha=0.05$}
\begin{adjustbox}{max width=\textwidth}
\begin{tabular}{lcccccccccc}
\toprule
& \multicolumn{2}{c}{\textbf{CIA (Split)}} & \multicolumn{2}{c}{\textbf{CIA (CQR)}} & \multicolumn{2}{c}{\textbf{CIA (Split) Stratified}} & \multicolumn{2}{c}{\textbf{CIA (CQR) Stratified}} & \multicolumn{2}{c}{\textbf{Group (Split)}} \\
\cmidrule(lr){2-3} \cmidrule(lr){4-5} \cmidrule(lr){6-7} \cmidrule(lr){8-9} \cmidrule(lr){10-11}
Dataset & Coverage & Size & Coverage & Size & Coverage & Size & Coverage & Size & Coverage & Size \\
\midrule

Anaheim & 0.94 (0.03) & 25.48 (0.85) & 0.94 (0.03) & 23.65 (1.08) & 0.94 (0.03) & 24.89 (0.98) & 0.94 (0.02) & 23.92 (1.10) & 0.88 (0.03) & 23.25 (0.90) \\
bike & 0.95 (0.02) & 6.52 (0.92) & 0.95 (0.02) & 6.04 (0.18) & 0.95 (0.02) & 5.24 (0.47) & 0.95 (0.02) & 1.09 (0.21) & 0.94 (0.02) & 4.25 (0.23) \\
Chicago & 0.95 (0.03) & 30.40 (1.06) & 0.95 (0.02) & 27.39 (1.46) & 0.95 (0.02) & 28.14 (1.53) & 0.95 (0.02) & 27.40 (1.53) & 0.84 (0.04) & 26.57 (1.36) \\
community & 0.95 (0.02) & 3.38 (0.33) & 0.95 (0.02) & 2.98 (0.26) & 0.95 (0.02) & 3.44 (0.24) & 0.95 (0.02) & 2.21 (0.26) & 0.94 (0.03) & 3.27 (0.45) \\
meps\_19 & 0.95 (0.02) & 42.86 (6.70) & 0.95 (0.02) & 44.27 (8.25) & 0.95 (0.02) & 49.28 (6.82) & 0.95 (0.02) & 45.37 (6.96) & 0.92 (0.02) & 40.41 (5.42) \\
meps\_20 & 0.95 (0.02) & 61.36 (10.31) & 0.95 (0.02) & 63.27 (8.80) & 0.95 (0.02) & 68.88 (8.32) & 0.95 (0.02) & 61.65 (7.37) & 0.92 (0.02) & 55.06 (6.05) \\
meps\_21 & 0.95 (0.02) & 40.78 (5.89) & 0.95 (0.02) & 44.86 (7.31) & 0.95 (0.02) & 49.85 (7.13) & 0.95 (0.02) & 46.45 (7.08) & 0.93 (0.02) & 42.24 (5.38) \\

\midrule
& \multicolumn{2}{c}{\textbf{Group (CQR)}} & \multicolumn{2}{c}{\textbf{Normal (Hetero)}} & \multicolumn{2}{c}{\textbf{Normal (Homo)}} & \multicolumn{2}{c}{\textbf{Bonferroni (Split)}} & \multicolumn{2}{c}{\textbf{Bonferroni (CQR)}} \\
\cmidrule(lr){2-3} \cmidrule(lr){4-5} \cmidrule(lr){6-7} \cmidrule(lr){8-9} \cmidrule(lr){10-11}
Dataset & Coverage & Size & Coverage & Size & Coverage & Size & Coverage & Size & Coverage & Size \\
\midrule

Anaheim & 0.92 (0.02) & 22.85 (0.91) & 0.86 (0.03) & 21.76 (0.84) & 0.90 (0.02) & 22.21 (0.82) & 0.99 (0.00) & 27.59 (1.30) & 0.99 (0.01) & 26.24 (1.16) \\
bike & 0.95 (0.02) & 5.06 (0.21) & 0.90 (0.02) & 5.21 (0.11) & 0.93 (0.01) & 3.79 (0.10) & 0.99 (0.01) & 28.90 (1.21) & 0.99 (0.00) & 26.89 (1.09) \\
Chicago & 0.86 (0.03) & 25.92 (1.39) & 0.82 (0.03) & 25.60 (1.34) & 0.84 (0.03) & 25.71 (1.33) & 1.00 (0.00) & 32.74 (2.02) & 0.99 (0.01) & 30.06 (1.81) \\
community & 0.94 (0.03) & 2.96 (0.36) & 0.81 (0.02) & 2.15 (0.03) & 0.92 (0.02) & 2.81 (0.14) & 0.95 (0.02) & 4.44 (0.33) & 0.95 (0.02) & 3.89 (0.27) \\
meps\_19 & 0.93 (0.02) & 43.71 (5.20) & 0.60 (0.02) & 7.66 (0.09) & 0.93 (0.01) & 33.38 (2.76) & 0.98 (0.01) & 389.89 (40.47) & 0.98 (0.01) & 389.91 (42.76) \\
meps\_20 & 0.93 (0.02) & 58.36 (6.39) & 0.60 (0.02) & 8.86 (0.10) & 0.93 (0.01) & 43.43 (3.09) & 0.98 (0.01) & 578.71 (54.66) & 0.98 (0.01) & 563.85 (50.91) \\
meps\_21 & 0.93 (0.02) & 43.32 (5.23) & 0.60 (0.02) & 7.33 (0.07) & 0.94 (0.01) & 35.73 (2.90) & 0.98 (0.01) & 400.18 (42.43) & 0.98 (0.01) & 370.19 (44.47) \\

\bottomrule
\end{tabular}
\end{adjustbox}
\end{table*}

\begin{table*}
\centering
\caption{Coverage and Size Results for Different Methods Across Datasets at $\alpha=0.01$}
\begin{adjustbox}{max width=\textwidth}
\begin{tabular}{lcccccccccc}
\toprule
& \multicolumn{2}{c}{\textbf{CIA (Split)}} & \multicolumn{2}{c}{\textbf{CIA (CQR)}} & \multicolumn{2}{c}{\textbf{CIA (Split) Stratified}} & \multicolumn{2}{c}{\textbf{CIA (CQR) Stratified}} & \multicolumn{2}{c}{\textbf{Group (Split)}} \\
\cmidrule(lr){2-3} \cmidrule(lr){4-5} \cmidrule(lr){6-7} \cmidrule(lr){8-9} \cmidrule(lr){10-11}
Dataset & Coverage & Size & Coverage & Size & Coverage & Size & Coverage & Size & Coverage & Size \\
\midrule

Anaheim & 0.99 (0.01) & 29.68 (1.95) & 0.99 (0.01) & 27.11 (1.61) & 0.98 (0.02) & 26.22 (1.04) & 0.98 (0.01) & 25.31 (1.06) & 0.95 (0.02) & 25.03 (1.08) \\
bike & 0.99 (0.01) & 13.83 (2.03) & 0.99 (0.01) & 7.40 (0.63) & 0.98 (0.01) & 7.04 (0.76) & 0.98 (0.01) & 2.18 (0.47) & 0.99 (0.01) & 6.63 (0.39) \\
Chicago & 0.99 (0.01) & 33.53 (1.14) & 0.99 (0.01) & 29.03 (1.49) & 0.99 (0.01) & 29.54 (1.57) & 0.99 (0.01) & 28.78 (1.54) & 0.95 (0.02) & 28.50 (1.58) \\
community & 0.99 (0.01) & 5.19 (0.60) & 0.99 (0.01) & 4.89 (1.07) & 0.99 (0.01) & 4.85 (0.37) & 0.99 (0.01) & 4.02 (0.79) & 0.99 (0.01) & 5.37 (0.77) \\
meps\_19 & 0.99 (0.01) & 101.43 (11.42) & 0.99 (0.01) & 102.07 (10.24) & 0.98 (0.01) & 104.16 (16.15) & 0.98 (0.01) & 97.98 (15.43) & 0.98 (0.01) & 88.53 (9.06) \\
meps\_20 & 0.99 (0.01) & 128.68 (16.50) & 0.99 (0.01) & 119.37 (14.50) & 0.98 (0.01) & 124.92 (16.46) & 0.98 (0.01) & 104.97 (11.72) & 0.98 (0.01) & 114.05 (9.08) \\
meps\_21 & 0.99 (0.01) & 104.26 (14.87) & 0.99 (0.01) & 101.38 (11.80) & 0.98 (0.01) & 99.60 (17.51) & 0.98 (0.01) & 90.37 (14.42) & 0.98 (0.01) & 94.13 (10.31) \\

\midrule
& \multicolumn{2}{c}{\textbf{Group (CQR)}} & \multicolumn{2}{c}{\textbf{Normal (Hetero)}} & \multicolumn{2}{c}{\textbf{Normal (Homo)}} & \multicolumn{2}{c}{\textbf{Bonferroni (Split)}} & \multicolumn{2}{c}{\textbf{Bonferroni (CQR)}} \\
\cmidrule(lr){2-3} \cmidrule(lr){4-5} \cmidrule(lr){6-7} \cmidrule(lr){8-9} \cmidrule(lr){10-11}
Dataset & Coverage & Size & Coverage & Size & Coverage & Size & Coverage & Size & Coverage & Size \\
\midrule

Anaheim & 0.97 (0.02) & 24.39 (1.01) & 0.93 (0.02) & 23.49 (0.88) & 0.96 (0.02) & 24.13 (0.86) & 1.00 (0.00) & 29.88 (1.78) & 0.99 (0.00) & 26.97 (1.20) \\
bike & 0.99 (0.01) & 7.22 (0.36) & 0.93 (0.02) & 6.85 (0.15) & 0.97 (0.01) & 4.98 (0.13) & 1.00 (0.00) & 33.97 (1.57) & 1.00 (0.00) & 31.50 (1.57) \\
Chicago & 0.95 (0.02) & 27.48 (1.48) & 0.93 (0.02) & 27.41 (1.39) & 0.95 (0.02) & 27.57 (1.39) & 1.00 (0.00) & 35.64 (2.38) & 0.99 (0.00) & 31.57 (1.87) \\
community & 0.99 (0.01) & 5.42 (1.30) & 0.86 (0.02) & 2.83 (0.04) & 0.96 (0.02) & 3.69 (0.18) & 0.99 (0.01) & 6.48 (0.50) & 0.99 (0.01) & 5.93 (0.89) \\
meps\_19 & 0.98 (0.01) & 92.53 (9.94) & 0.69 (0.02) & 10.06 (0.12) & 0.95 (0.01) & 43.87 (3.63) & 0.99 (0.01) & 662.98 (68.44) & 0.99 (0.01) & 665.74 (62.79) \\
meps\_20 & 0.98 (0.01) & 113.42 (9.68) & 0.68 (0.02) & 11.65 (0.13) & 0.95 (0.01) & 57.07 (4.06) & 0.99 (0.01) & 892.21 (54.55) & 0.99 (0.01) & 856.86 (52.43) \\
meps\_21 & 0.98 (0.01) & 92.66 (10.70) & 0.70 (0.02) & 9.64 (0.09) & 0.95 (0.01) & 46.96 (3.81) & 0.99 (0.01) & 682.08 (59.95) & 0.99 (0.01) & 653.63 (58.24) \\

\bottomrule
\end{tabular}
\end{adjustbox}
\end{table*}

\end{document}